\pdfoutput=1

\documentclass[11pt]{article}

\usepackage[final]{acl}

\usepackage{times}
\usepackage{latexsym}

\usepackage[T1]{fontenc}

\usepackage[utf8]{inputenc}

\usepackage{microtype}

\usepackage{inconsolata}

\usepackage{graphicx}

\usepackage{xcolor}
\usepackage{array}
\usepackage{tabularx}
\usepackage{amsmath}
\usepackage[vlined,ruled]{algorithm2e}
\usepackage{array,ragged2e,enumitem}

\newcommand{\mathsc}[1]{\text{\textsc{#1}}}
\definecolor{myorange}{rgb}{1.0, 0.5, 0.0}
\definecolor{myred}{rgb}{0.8, 0.0, 0.0}
\definecolor{mildgreen}{rgb}{0.0, 0.5, 0.0}

\definecolor{nicegreen}{RGB}{192,239,171}
\definecolor{niceblue}{RGB}{174,224,246}
\definecolor{nicered}{RGB}{239,171,192}
\usepackage{cleveref}
\usepackage{multirow}
\usepackage{multicol}
\usepackage{hyperref}

\title{LiteraryQA: Towards Effective Evaluation of Long-document\\ Narrative QA}

\author{Tommaso Bonomo\textsuperscript{1}\thanks{Equal contribution.}, Luca Gioffré\textsuperscript{1}\footnotemark[1], 
 \normalfont{and} {\bf Roberto Navigli\textsuperscript{1,2}}\\\\
\textsuperscript{1}Sapienza NLP Group, Sapienza University of Rome \quad \textsuperscript{2}Babelscape, Italy \\
\texttt{\{bonomo, gioffre, navigli\}@diag.uniroma1.it}
}

\begin{document}
\maketitle

\begin{abstract}

Question Answering (QA) on narrative text poses a unique challenge to current systems, requiring a deep understanding of long, complex documents. However, the reliability of NarrativeQA, the most widely used benchmark in this domain, is hindered by noisy documents and flawed QA pairs.
In this work, we introduce LiteraryQA, a high-quality subset of NarrativeQA focused on literary works. Using a human- and LLM-validated pipeline, we identify and correct low-quality QA samples while removing extraneous text from source documents. 
We then carry out a meta-evaluation of automatic metrics to clarify how systems should be evaluated on LiteraryQA.
This analysis reveals that all \textit{n}-gram-based metrics have a low system-level correlation to human judgment, while LLM-as-a-Judge evaluations, even with small open-weight models, can strongly agree with the ranking identified by humans.
Finally, we benchmark a set of long-context LLMs on LiteraryQA.
We release our code and data at \href{https://github.com/sapienzaNLP/literaryQA}{github.com/SapienzaNLP/LiteraryQA}.
\end{abstract}

\section{Introduction}

Question Answering (QA) has long been a core task in Natural Language Processing, supported by a large number of datasets that differ one from another across several dimensions~\citep{rogers-etal-2023-qa}: question type and objective, answer format, and given context.
These datasets have enjoyed widespread adoption by the community, making up an important part of the evaluations of current models~\citep{olmo20242olmo2furious,anthropic_claud37_sonnet,qwen2.5,deepseekai2025deepseekr1incentivizingreasoningcapability}.
A particular QA setting is the one that focuses on whole books and narrative corpora.
Books, and in general narrative text, express intricate sequences of events that unfold across very long text, as outlined by characters or by an external narrator~\cite{piper-etal-2021-narrative}.
Even the latest Large Language Models (LLMs) find this setting challenging, as they have to understand the underlying plot and link information from different parts of the (long) story~\cite{pang-etal-2022-quality,wang2025novelqa}.

\begin{figure}[t]
    \centering
    \includegraphics[width=\linewidth]{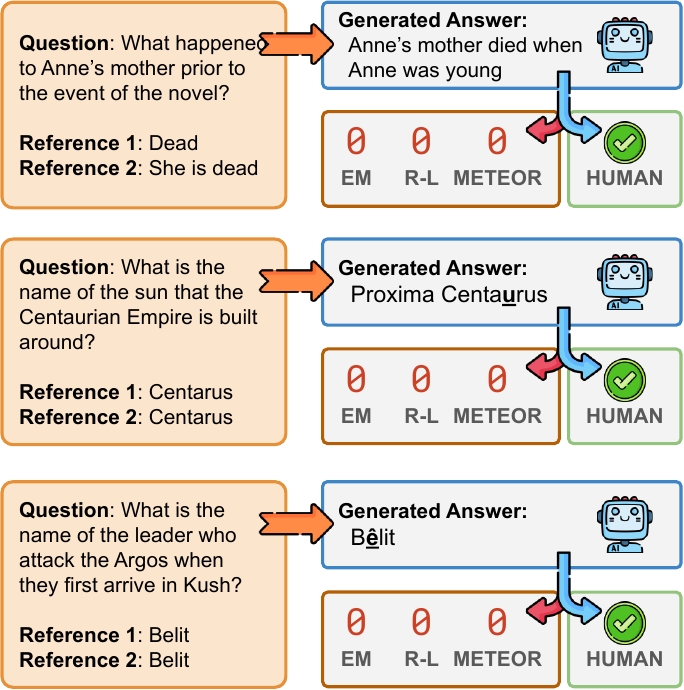}
    \caption{Illustrative example of the failure modes that automatic metrics incur when evaluating predictions on the original NarrativeQA.}
    \label{fig:example}
\end{figure}


In this context, NarrativeQA~\citep{kocisky-etal-2018-narrativeqa} is arguably the most established benchmark for the evaluation of long-context models' abstractive QA capabilities on English narrative text.
It is included in many long-context benchmarks, notably \(\infty\)Bench\citep{zhang-etal-2024-bench}, L-Eval~\citep{an-etal-2024-l}, LongBench~\citep{bai-etal-2024-longbench,bai-etal-2025-longbench}, and HELMET~\citep{yen2025helmet}.
NarrativeQA is very different from other free-form QA datasets, as a majority of its questions require understanding and differentiating narrative events and their relations~\citep{mou-etal-2021-narrative}.
As we show in our work, NarrativeQA also contains \textit{noisy} content: there are instances of misaligned summaries and source texts, questions and answers that are grammatically and semantically incorrect with respect to the reference summary, and incorrect or malformed reference answers.


Moreover, model performance on NarrativeQA is comparatively low: it is unclear if the underlying reason stems from the inherent difficulty of abstractive questions on narrative text, the flaws present in the dataset, or the unsuitable metrics used to evaluate predictions with respect to reference answers.
Many automatic metrics have been used to evaluate performance on NarrativeQA: \citet{kocisky-etal-2018-narrativeqa} measured BLEU-1, BLEU-4~\citep{papineni-etal-2002-bleu}, METEOR~\citep{banerjee-lavie-2005-meteor} and ROUGE-L~\citep{lin-2004-rouge}, while later evaluations adopted token-level F1 from extractive QA tasks~\citep{shaham-etal-2022-scrolls,an-etal-2024-l} or tasked a model to evaluate if an answer is correct or not, a paradigm referred to as LLM-as-a-judge~\citep{chen-etal-2019-evaluating, wang_evaluating_2023}.
Except for the latter, these \textit{n}-gram-based metrics rely on an exact match between the words appearing in the reference answers and in the prediction of a system.
Figure~\ref{fig:example} exemplifies this kind of issue.
These metrics can assign low scores, or even zero, to semantically correct responses that contain minor typos, missing diacritics, or are paraphrases of the reference answers.
In contrast, human readers would easily recognize the correctness of such responses beyond surface-level variations.
To the best of our knowledge, there has not been a comprehensive study to ascertain which automatic metric is most correlated with human judgment in the context of abstractive, narrative QA pairs.


In order to address these issues, we propose LiteraryQA, a human- and LLM-validated subset of NarrativeQA focused only on literary works.
Following recent literature that found LLMs to be capable annotators~\citep{Gilardi2023ChatGPTOC,pmlr-v239-mohta23a}, we employ Claude 3.5 Haiku~\citep{anthropic_claude_haiku} in a multi-step pipeline to first identify, and subsequently correct, questions and answers that are not acceptable according to a set of guidelines.
We also carry out an extensive meta-analysis on which automatic metric to use, according to its agreement with human judgment, considering common \textit{n}-gram-based metrics and LLM-as-a-judge solutions.
We then benchmark current long-context LLMs, both open- and closed-weights, on both NarrativeQA and LiteraryQA, demonstrating the challenge these types of datasets pose to current state-of-the-art systems.

\section{Related Work}

\subsection{Narrative-based QA}
NarrativeQA~\cite{kocisky-etal-2018-narrativeqa} was an early effort to scale QA to entire books and movie scripts, with an average length of around 60,000 tokens. Despite its scale and free-form format, answers tend to be short and often paraphrased from summaries, leading to the inconsistencies pointed out in the Introduction.
Recent benchmarks have advanced long-context QA over narrative texts. 
QuALITY~\cite{pang-etal-2022-quality} presents multiple-choice questions over medium-length fiction text, averaging 5,159 tokens, which cannot be considered long-context in modern scenarios.
NarrativeXL~\cite{moskvichev-mai-2023-narrativexl} scales to 700k multiple-choice questions across 1,500 novels, but its reliance on structured questions limits its semantic depth.
Contemporarily to our work, NovelQA~\cite{wang2025novelqa} offers full-book contexts and a choice between multiple possible answers to a question, although it is restricted to 60 publicly available books.
Unfortunately, to avoid data leakage, they do not publicly release the correct answers or the pieces of evidence required to correctly answer a question.
We include a full analysis of the differences between our work and NovelQA in Appendix \cref{app:novelqa}.


\subsection{Long-document Resources}
Various resources have annotated narrative text for tasks besides QA: LitBank~\citep{bamman-etal-2019-annotated,bamman-etal-2020-annotated,sims-etal-2019-literary} provides annotations for Literary Event Detection, Literary Entities and Coreference Resolution on the first 2000 tokens of 100 literary works, while \textsc{BookCoref}~\citep{martinelli-etal-2025-bookcoref} provides silver- and gold-quality Coreference Resolution data on the full text of 53 books.
Beyond narrative text, several benchmarks target long-document reasoning across diverse domains. The SCROLLS benchmark suite \cite{shaham-etal-2022-scrolls} aggregates tasks such as summarization and QA over government reports \cite[GovReport]{huang-etal-2021-efficient}, TV transcripts \cite[SummScreenFD]{chen-etal-2022-summscreen}, and meeting notes \cite[QMSum]{zhong-etal-2021-qmsum}. While valuable for studying long-context understanding, it does not investigate the specific narrative setting that we are interested in.
Other QA-specific datasets include Qasper \cite{dasigi-etal-2021-dataset}, which requires fine-grained fact retrieval from research papers, and ContractNLI \cite{koreeda-manning-2021-contractnli-dataset}, which frames contract understanding as a document-level entailment task.
To probe deep retrieval and reasoning, RULER \cite{hsieh2024ruler} introduces synthetic tasks over extremely long sequences, such as variable tracking and information chaining. While they are useful for stress-testing model capacities, synthetic tasks may not reflect the complexity of real-world documents.

Our work contributes a natural, generative QA benchmark over long-form narrative texts, designed to balance document-level scope, high-quality supervision, and flexible answer generation.

\subsection{Meta-evaluation of QA Metrics}


Many works have carried out a meta-evaluation of metrics for free-form, generative QA.
\citet{kamalloo-etal-2023-evaluating} and \citet{wang_evaluating_2023} focused on factual, Wikipedia-based QA datasets, reporting contrasting findings regarding the correlation of metrics based on lexical overlap with human judgments.
Instead, \citet{chen-etal-2019-evaluating} carry out a study in the narrative domain, where they find a moderately high correlation for \textit{n}-gram- and neural-based metrics. However, their study analyses predictions from older models equipped with a copying mechanism, which makes them lexically similar to the references. Moreover, using the summary instead of the full text of a book
makes the setting significantly easier than long-document narrative QA.

To the best of our knowledge, no prior work has conducted a comprehensive meta-evaluation of QA metrics in the narrative domain.
We address this gap by providing the first systematic analysis of how standard evaluation metrics perform when applied to narrative question answering.

\section{LiteraryQA \label{sec:literaryqa}}
We hypothesize that the challenging aspect of NarrativeQA can be partly ascribed to inconsistencies in text quality and formatting (including HTML artifacts and unrelated content), and to problematic QA samples containing wrong and misspelled reference answers or unanswerable questions, which could artificially lower the performance of the systems
To mitigate these problems, we develop a human-curated data refinement pipeline that, applied to NarrativeQA, creates an improved high-quality dataset, LiteraryQA.
Since our main goal is to provide a benchmark for narrative QA, we manually validate only the test set. We run the whole pipeline also on the train and validation sets in order to release the full dataset.

\subsection{Data Refinement Pipeline}
Our pipeline is composed of two main phases: document-level and QA-level.
In the following sections, we detail our filtering approach designed to produce a more balanced and narrative-representative dataset. 

\subsubsection{Document-level Phase}
Our preliminary qualitative inspection of NarrativeQA reveals potential concerns regarding the pairing between book texts and their corresponding summaries, raising the need for a systematic alignment check. 
This is a fundamental issue since documents with misaligned summaries will have unanswerable questions, as the QA samples cannot be answered from an unrelated source text.
Moreover, NarrativeQA contains different document types, spanning novels, movie screenplays, poetry collections, theatrical plays, fairytales, and other types of text that may not strictly fit the conventional narrative text definition~\cite{piper-etal-2021-narrative}. 
This heterogeneity introduces substantial variance in the dataset in terms of format and style, and distracts from the challenge of understanding a narrative plot. Thus, we limit our focus to the book categories in order to develop a structurally and stylistically homogeneous narrative dataset. 

\paragraph{Document Filtering}
First, we manually annotate all documents in the book category\footnote{Movie documents are clearly categorized in NarrativeQA, making it possible to filter them out easily.} of the test set so as to identify and exclude mismatched documents, theatrical plays, and non-narrative texts.
Our annotation process reveals that, out of the 177 books in the test set, there are 8 mismatched samples ($4.5\%$), 20 theatrical plays ($11.3\%$), and 11 non-narrative documents ($6.2\%$), for a total of 39 documents ($22\%$) that we subsequently remove from the dataset, resulting in 138 documents kept in the filtered test set.
In addition to the manual annotations, we also develop an automatic approach employing an LLM to classify the training and validation sets' samples. 
For each document, we prompt an LLM either with the Wikipedia page of the document or with the starting paragraphs of the text (\Cref{tab:categorizer_prompt_play,tab:categorizer_prompt_non_fiction,tab:categorizer_prompt_mismatched}) and validate this approach on the test set.

\paragraph{Text Cleaning}
When examining the filtered documents, we discovered that many documents contain text unrelated to the book, which inflates document length and could confuse models. 
Such text included HTML and Markdown strings, Project Gutenberg headers and footers, and legal license sections.
To address this issue, we downloaded the original HTML versions of all documents through the URLs included in the dataset\footnote{We used Project Gutenberg's mirrors as some of the original URLs are no longer available.}.
Then, we devised an algorithm to isolate the narrative content of each book through a set of heuristics. This algorithm was iteratively refined on the manually cleaned test set, with each iteration undergoing manual validation.
Throughout this process, we prioritized recall over precision, ensuring that all narrative elements were preserved, even at the cost of including occasional non-narrative content. 
More details can be found in \Cref{app:pipeline_details}. 

We also fixed several encoding errors within the summaries, particularly regarding incorrect diacritical marks (e.g., \texttt{Ăvariste} instead of \texttt{Évariste}).
On average, our cleaning procedure produced documents of 3K tokens shorter than the original texts in NarrativeQA (\Cref{fig:token_count_LQAvsNQA} in \Cref{app:LQA_stats}).

\subsubsection{QA-level Steps}
Our second refinement phase focuses on individual question-answer pairs.
Upon manual inspection, we found duplicated questions within the same book, grammatical errors, and issues with the semantic correctness of question-answer pairs.
Given the high number (4223) of QA samples in the filtered NarrativeQA test split, we employed an LLM to identify and correct QA issues, validating its outputs on a set of 20 documents spanning different genres and authors.
At the end of the pipeline, 1608 samples ($38\%$) were modified. 
Appendix~\ref{app:LQA_stats} lists these annotated documents, along with additional statistics on the test set and the full guidelines and prompts for QA refinement.

\paragraph{Question Deduplication}
We implemented a simple ROUGE filtering mechanism in which we identify and remove 125 (1.2\%) duplicate questions (i.e., questions exceeding a ROUGE-L similarity threshold) after manual validation\footnote{We repeated this step at the end of the pipeline to remove new duplicates introduced with the LLM corrections.}.

\paragraph{Questions Refinement}
After deduplication, we wanted to assess whether the questions were acceptable from both a grammatical and semantic point of view.
We defined \textit{malformed questions} as questions containing lexical errors, such as misspelled character names or typographical mistakes, as well as grammatical issues. 
\textit{Ill-posed questions}, on the other hand, were defined as questions containing false assumptions, misrepresenting facts presented in the summary, or being fundamentally unanswerable based on the available information (\Cref{tab:annotation_guidelines}, \Cref{app:LQA_stats}).

We tasked an LLM with identifying and correcting malformed and ill-posed questions. 
Since the original questions of NarrativeQA were generated from the summaries, we provided the summary as a reference to the LLM (\Cref{tab:prompt-questions}, \Cref{app:LQA_stats}).

\paragraph{Answers Refinement}
We evaluated the reference answers following a similar approach. 
We applied the exact criteria used for malformed questions to identify \textit{malformed answers}, and we defined \textit{invalid answers} as answers failing to be either i) factually accurate, ii) complete in addressing all aspects of the question, or iii) directly relevant to the information requested (\Cref{tab:annotation_guidelines}, \Cref{app:LQA_stats}).

As in the previous step, we used an LLM to identify and correct these issues. 
We prompted it with the document summary, the question that had passed our previous refinement step (either because originally correct or subsequently corrected) and the reference answer to evaluate\footnote{We evaluated each reference answer independently.} (\Cref{tab:prompt-answers}, \Cref{app:LQA_stats}).


\subsection{Evaluation of Pipeline Steps}

When designing our pipeline, we prioritize precision over recall to ensure that only high-quality samples contribute to the final dataset. 
Each step in the pipeline processes the output from the previous step, creating a cascading refinement process.

Regarding document-filtering step for the training and validation sets, we use a Llama 3.1 8B Instruct  model~\citep{grattafiori2024llama3herdmodels}. 
We validate the model classification outputs on the test set, resulting in good performance (\Cref{tab:doc_pipeline_performance}, \Cref{app:LQA_stats}).
We believe this automated approach offers a promising foundation for addressing issues in the training and validation datasets. 
However, we defer this work to future research due to the prohibitive scale of manual refinement required and focus only on the test set for the last part of the evaluation.

For the QA-level steps, which pose a greater evaluation challenge compared to the document-level phase, we employ Claude 3.5 Haiku \citep{anthropic_claude_haiku}. 
We evaluate the quality of the QA-level steps of our pipeline by examining and annotating the outputs of the LLM.
Except for the question deduplication step, for which all identified duplicates are examined, we perform our evaluation on a selected subset of 20 documents from the test set, comprising a total of 583 QA samples ($15\%$).
These documents are listed in \Cref{tab:annotated_docs}, \Cref{app:LQA_stats}.
Two of the authors performed the annotations necessary to assess pipeline quality, analyzing the question subset described above.
The annotation process required approximately 30 hours total for each annotator.

Initially, annotators validate the original questions and two reference answers according to the criteria established in our methodology (\Cref{tab:annotation_guidelines}, \Cref{app:LQA_stats}).
Inter-annotator agreement was measured using Cohen's Kappa coefficient, yielding an average \(\kappa = 0.83\), which indicates excellent agreement (\Cref{tab:iaa_qa_samples}).

\begin{table}[t]
\begin{tabularx}{\columnwidth}{Xccc}
\hline
Acceptability & $\kappa$ & A1 \% & A2 \% \\ \hline
Questions     & 0.75   & 80.24      & 85.25 \\
Answer \#1    & 0.71   & 86.60      & 84.54 \\
Answer \#2    & 0.68   & 81.96      & 74.70 \\ \hline
Average       & 0.71   & 82.93      & 81.50 \\ \hline
\end{tabularx}
\caption{Inter-annotator agreement on the classification of 583 QA samples (20 documents) in the original NarrativeQA test set, before refinement. Values in columns A1 and A2 are the percentage of accepted modifications according to the respective annotators.}
\label{tab:iaa_qa_samples}
\end{table}

\begin{table}[t]
\begin{tabularx}{\columnwidth}{Xccc}
\hline
Corrections & $\kappa$ & Acc. A1 & Acc. A2 \\
\hline
Only in Questions & $0.88$ & $0.95$  & $0.96$ \\
Only Answers   & $1.00$   & $0.96$  & $0.96$ \\
Both      & $1.00$   & $0.65$  & $0.65$ \\
\hline     
\end{tabularx}
\caption{Analysis on 176 QA samples of the annotated subset modified by Claude 3.5 Haiku.
We report the accuracy of the corrections based on the judgments of the two annotators (A1 and A2),  and the inter-annotator agreement with Cohen's Kappa.}
\label{tab:validness_check_claude_corrections}
\end{table}

Then, to assess the quality of the LLM corrections, we evaluated the samples of the annotated subset modified by the LLM. 
%
This qualitative analysis reveals that many false positives (instances classified as acceptable by human annotators but rejected and subsequently corrected by the LLM) involved only minor modifications. 
These corrections typically produce paraphrases that preserve the essential meaning of the original samples.


We also observe that samples with corrections to either the question or the answer result in predominantly correct instances. 
However, samples with corrections to both the question and answer often contain compounding errors that render the QA pairs invalid. 
Based on this finding, we exclude all QA samples with double corrections from our dataset.
We present the quantitative results of this last analysis on the 176 modified samples of the annotated subset in \Cref{tab:validness_check_claude_corrections} and some examples in \Cref{tab:claude_corrections}, \Cref{app:LQA_stats}.

\Cref{tab:document_filtering} presents an overall breakdown of the impact of each step of our pipeline on the test set of NarrativeQA.
We also show the length distribution of question-answer pairs before and after processing in \Cref{tab:qa_stats}. 
While modified answers show greater length variability, the mean length remains consistent across both versions.


\begin{table}[t]
\centering
\begin{tabularx}{\columnwidth}{X>{\raggedleft\arraybackslash}r>{\raggedleft\arraybackslash}r}
\hline
Step      & \# Docs & \# QAs \\ \hline
NarrativeQA (original)        & $355$   & $10557$ \\
$-\ $Movies        & $-178$  & $-5207$ \\
$-\ $Plays         & $-20$   & $-573$  \\
$-\ $Other         & $-11$   & $-234$  \\ 
$-\ $Mismatched    & $-8$    & $-320$  \\ \hline
NarrativeQA (filtered)           & $138$   & $4223$  \\  
$-\ $QA duplicates & -       & $-125$  \\
$-\ $Double Correction & -       & $-308$  \\ 
$-\ $QA duplicates & -       & $-5$    \\ \hline
LiteraryQA& $138$   & $3785$  \\ 
\hline
\end{tabularx}
\caption{Breakdown of the impact of our data refinement pipeline on the LiteraryQA test set. The first document-level phase produces a `filtered' version of NarrativeQA, whilst the QA-level refinement yields the final LiteraryQA test set (last row). The `QA duplicates' step is run twice to remove duplicates introduced by the LLM, and `Double Correction' stands for the samples where the LLM modified both the question and the answer, which we chose to exclude.}
\label{tab:document_filtering}
\end{table}

\begin{table}[t]
\begin{tabularx}{\columnwidth}{l>{\centering\arraybackslash}X>{\centering\arraybackslash}X}
\hline
Length in tokens               & $\mu$  & $\sigma$ \\ \hline
NarrativeQA questions          & $8.60$ & $\pm3.30$ \\
LiteraryQA questions  & $8.62$ & $\pm3.24$ \\
Only modified questions      & $9.76$ & $\pm3.43$ \\
\hline
NarrativeQA answers            & $4.22$ & $\pm3.63$ \\
LiteraryQA answers    & $4.33$ & $\pm4.07$ \\ 
Only modified answers        & $6.86$ & $\pm6.21$ \\
\hline
\end{tabularx}
\caption{Average length of questions and answers in the original NarrativeQA dataset compared to LiteraryQA after applying our Data Refinement pipeline. We also report statistics for only the modified samples.}
\label{tab:qa_stats}
\end{table}

\newpage

\section{Metrics Analysis}
We analyze several metrics on LiteraryQA, ranging from traditional \textit{n}-gram-based approaches to neural-based methods, to measure their system-level correlation with human annotations on LiteraryQA. 
This step is necessary to establish the most suited evaluation metric for QA on narrative text, as previous approaches have simply adopted existing metrics from other QA domains.
Following recent work that confirmed LLMs to be capable evaluators~\citep{li2024llmasajudge,gu2024asurvey}, we also include LLM-as-a-judge as a metric to compare and contrast its agreement with \textit{n}-gram- and neural-based measures.
The LLM-as-a-judge paradigm involves querying an LLM with a question, multiple reference answers, a context, and a candidate answer, obtaining a score that the LLM generates according to a rubric provided through system instructions.
By measuring the system-level correlation between a metric and human judgment, we assess how closely the metric's ranking 
aligns with the human preferences.

We also compare the differences in system-level correlation of \textit{n}-gram-based and neural-based metrics and LLM-as-a-judge approaches on LiteraryQA against NarrativeQA: if a metric on the former correlates better with human judgment than the same metric on the latter, it would indicate that a large amount of noise from the dataset has been captured and corrected by our pipeline.

We include metrics that have been used in literature to evaluate answers on NarrativeQA, namely: ROUGE-L~\citep{lin-2004-rouge}, METEOR~\citep{banerjee-lavie-2005-meteor}, token-level F1 (F1) and exact-match (EM) taken from extractive QA~\citep{yang-etal-2018-hotpotqa}.
As our neural-based metric, we use BERTScore~\citep{zhang2020bertscore}, which provides a score between 0 and 1 that represents the semantical similarity of two pieces of text.
Regarding the LLM-as-a-judge setup, we use state-of-the-art LLMs accessed through APIs (GPT 4.1\footnote{\texttt{gpt-4.1-2025-04-14}} and Claude 3.7 Sonnet\footnote{\texttt{claude-3-7-sonnet-20250219}}) and Prometheus 2 7B~\cite{kim-etal-2024-prometheus}, an evaluator LM finetuned to provide direct assessments of candidate answer quality according to a user-defined rubric.
Our benchmark requires models to process entire books for each question, so our computational budget\footnote{A single node equipped with 4 NVIDIA A100 GPUs.} limited our LLM selection to the above models.

\begin{table}[t]
\begin{tabularx}{\columnwidth}{Xcc}
\hline
Model                 & Size & Context \\
\hline
Qwen 2.5 \small\citep{yang2025qwen251mtechnicalreport}              & 7B  & 1M \\
Qwen 2.5 \small\citep{yang2025qwen251mtechnicalreport}              & 14B & 1M   \\
Llama 3.1 \small\citep{grattafiori2024llama3herdmodels}             & 8B  & 128K \\
NExtLong \small\citep{gao2025nextlongeffectivelongcontexttraining}  & 8B  & 512K \\
GLM-4 \small\citep{glm2024chatglm}                                  & 9B  & 1M   \\
\hline \\[-8pt]
\parbox[c]{\hsize}{Claude 3.5 Haiku \newline \small\citep{anthropic_claude_haiku}}               & ?   & 200K  \\[8pt]
\parbox[c]{\hsize}{Gemini 2.0 Flash Lite \\ \small\citep{google_gemini20_flash}}           & ?   & 1M    \\[8pt]
\hline
\end{tabularx}
\caption{Instruction-finetuned models tested, categorized as open-weight (first 5) or API-based (last 2). Inputs exceeding the context window are truncated.}
\label{tab:models_stats}
\end{table}

\subsection{Experimental Setup}

\paragraph{Human judgments}
We collect human judgments on the quality of generated responses.
We randomly sample \(N = 500\) QA pairs from LiteraryQA's test split and input the question to each of the \(M = 7\) systems in \Cref{tab:models_stats}, obtaining their prediction.
We repeat the same process for NarrativeQA, resulting in 7000 (QA, prediction) pairs collected across the two datasets.
Annotators then evaluate the quality of each prediction according to the rubric in \Cref{tab:grading_rubric}, \Cref{app:LQA_stats}. 
Given a question, its two reference answers, and an answer produced by a system, each annotator is required to score the automatic answer in a range between 1 and 5 in this \textbf{reference-based setting} following the rubric.
Annotators also evaluate each automatic answer in a separate \textbf{summary-based setting}, where they have access to the book summary as additional context. 

Two of the authors of this paper annotated 7000 predictions each, with an estimated annotation time of 50 hours across multiple sessions.
The inter-annotator agreement between them, measured through Kendall's \(\tau\) correlation, is 0.7876 for LiteraryQA and 0.8098 for NarrativeQA.


\begin{table*}[t]
\centering
\begin{tabularx}{\linewidth}{X|cc|cc}
\hline
\multirow{2}{*}{\large Metric} & \multicolumn{2}{c|}{\large NarrativeQA} & \multicolumn{2}{c}{\large LiteraryQA} \\
 & Reference-based & Summary-based & Reference-based & Summary-based \\
\hline
EM        & ~0.0325  \small[-0.48, 0.48] & - \quad\quad\quad -   & ~0.0614  \small [-0.24, 0.43] & - \quad\quad\quad -\\
F1        & ~0.0328   \small [-0.48, 0.48] & - \quad\quad\quad - & ~0.0574  \small [-0.24, 0.39] & - \quad\quad\quad -\\
ROUGE-L   & ~0.0291   \small [-0.48, 0.42] & - \quad\quad\quad - & ~0.0580  \small [-0.24, 0.39] & - \quad\quad\quad -\\
METEOR    & ~\underline{0.1519}   \small [-0.33, 0.62] & - \quad\quad\quad - & ~\underline{0.4444}  \small [~0.14, 0.81] & - \quad\quad\quad -\\
\hline
BERTScore & -0.0477  \small [-0.52, 0.39] & - \quad\quad\quad - & ~0.0677  \small [-0.24, 0.43] & - \quad\quad\quad -\\
\hline
Prometheus 2 7B   & ~0.2195  \small [-0.24, 0.68] & 0.3155  \small [-0.05, 0.68] & ~\textbf{0.4499}  \small [-0.05, 0.88] & \textbf{0.6881}  \small [~0.33, 0.98]  \\
Sonnet 3.7 & ~0.3114  \small [-0.14, 0.78] & \textbf{0.5829}  \small [~0.20, 0.90] & ~0.3651  \small [-0.14, 0.81] & 0.5243  \small [~0.09, 0.90]  \\
GPT 4.1 & \textbf{~0.3517}  \small [-0.10, 0.81] & 0.5080  \small [~0.09, 0.90] & ~0.3282  \small [-0.14, 0.71] & 0.5593  \small [~0.20, 0.90]  \\
\hline
\end{tabularx}
\caption{System-level Kendall's \(\tau\) correlation with human judgments on NarrativeQA and LiteraryQA. Bold and underline mark the best LLM-as-a-Judge and n-gram metrics, respectively. 95\% confidence intervals are in brackets. The summary-based setting is exclusive to the LLM-as-a-Judge method.}
\label{tab:correlations}
\end{table*}

\paragraph{Correlation measurement}
We compute the system-level correlation (\(r\)) to see how well a metric's ranking of systems aligns with human judgments. The calculation uses outputs from M systems on N documents, following the notation of~\citet{deutsch-etal-2021-statistical,deutsch-etal-2022-examining}.
Specifically,

\begin{gather*}
    r = \mathsc{Corr}\left(\left\{\left(\frac{1}{N}\sum_{j = 1}^{N} x^j_i, \frac{1}{N}\sum_{j = 1}^{N} z^j_i\right)\right\}_{i=1}^M\right)
\end{gather*}
where \(x_i^j\) and \(z_i^j\) are the scores assigned by the metric \(\mathcal{X}\) and human judgment \(\mathcal{Z}\), respectively, to the output of the \(i\)-th system on the \(j\)-th item, and $\mathsc{Corr}$ can be any measure of correlation, in our case Kendall's \(\tau\).\footnote{We chose Kendall's \(\tau\) as we want to measure the correlation in ranking power of a metric compared to human scores. We computed it through its implementation in \texttt{scipy}.}

Regarding \textit{n}-gram-based metrics, we calculate the correlation of EM, F1, ROUGE-L and METEOR on two sets of \(N \cdot M = 3500\) samples, one from LiteraryQA and one from NarrativeQA, which are our human-annotated judgments.
As a neural-based metric, we calculate the correlation of BERTScore equipped with DeBERTa-XLarge~\cite{he2021deberta} finetuned for NLI.

For the LLM-as-a-judge paradigm, we evaluate three LLMs to see how closely they correlate with human judgment, specifically GPT 4.1, Claude 3.7 Sonnet, and an open-weight option, Prometheus 2 7B.
All models are initialized with a system prompt that describes the annotation required and provides an evaluation rubric (the prompt follows the same rubric defined in Table~\ref{tab:grading_rubric}, \Cref{app:LQA_stats}).
Contrary to \textit{n}-gram-based metrics, LLMs can also incorporate extra context when assigning the score of a predicted answer.
We make use of this characteristic, as we did during the annotation process, and devise two settings in which we measure the system-level LLM-as-a-judge correlation:
\textbf{reference-based}, where the LLM is given only the question, the reference answers, and the candidate answer; and \textbf{summary-based}, where we also provide the model with the summary of the book, allowing it to disregard the reference answers when scoring a prediction if it can support it through the summary.

\begin{table*}[t]
\centering
\begin{tabularx}{\linewidth}{Xccccc|c|c}
\hline
Model & Context & R-L & METEOR & EM & F1 & BERTScore & Prometheus 2\\
\hline
Llama3.1-8B & 128K & 0.3904 & 0.3669 & 0.1663 & 0.3785 & 0.7105 & 2.981 \\
NExtLong-8B & 512K  & \textbf{0.4155} & 0.3617 & \textbf{0.2015} & \textbf{0.4057} & \textbf{0.7195} & 2.836\\
Qwen2.5-7B & 1M  & 0.3123 & 0.3311 & 0.0529 & 0.3033 & 0.6689 & 2.843\\
GLM-4-9B & 1M  & 0.3372 & \textbf{0.3849} & 0.0924 & 0.3319 & 0.6705 & 3.149\\
Qwen2.5-14B & 1M  & 0.3300 & 0.3632 & 0.0679 & 0.3216 & 0.6764 & 3.123\\
\hline
Claude 3.5 Haiku & 200K & 0.2534 & 0.2988 & 0.0425 & 0.2818 & 0.6569 & \textbf{3.296}\\
Gemini2-Flash-L & 1M  & 0.2299 & 0.2825 & 0.0158 & 0.2574 & 0.6440 & 2.860\\
\hline
\end{tabularx}
\caption{Performance of seven open-weight and closed-source models on LiteraryQA using four automatic \textit{n}-gram-based metrics, a neural-based metric (BERTScore) and an LLM-as-a-Judge (Prometheus 2). Best scores are in bold.}

\label{tab:lqa-performance}
\end{table*}

\subsection{Results}
\Cref{tab:correlations} presents the system-level correlations between our chosen metrics and human judgment, measured using Kendall's $\tau$. 
The metrics include four \textit{n}-gram-based measures, one neural-based metric, and three LLMs used as judges. 
In this section, we examine each category of metrics to discuss the results.
\paragraph{\textit{N}-gram-based metrics}
The results of \textit{n}-gram-based metrics show poor correlations in general, especially in NarrativeQA: except for METEOR, all other metrics are poorly correlated with our collected human judgments.
This reflects the fragility that these metrics demonstrate concerning noise in the reference answers.
Regarding METEOR, we hypothesize that its stemming and synonym-resolution features mitigate much of the noise that can be encountered in the original NarrativeQA.

On LiteraryQA, instead, \textit{n}-gram metrics (except METEOR) have a slightly better correlation with human judgment, which indicates that it contains questions and reference answers of higher quality compared to NarrativeQA.
This demonstrates the effectiveness of the pipeline we showcase in Section~\ref{sec:literaryqa}.
METEOR actually achieves a good system-level correlation with human judgment of 0.44, indicating that it should be preferred among all \textit{n}-gram-based metrics.

\begin{figure}[t]
    \includegraphics[width=\columnwidth]{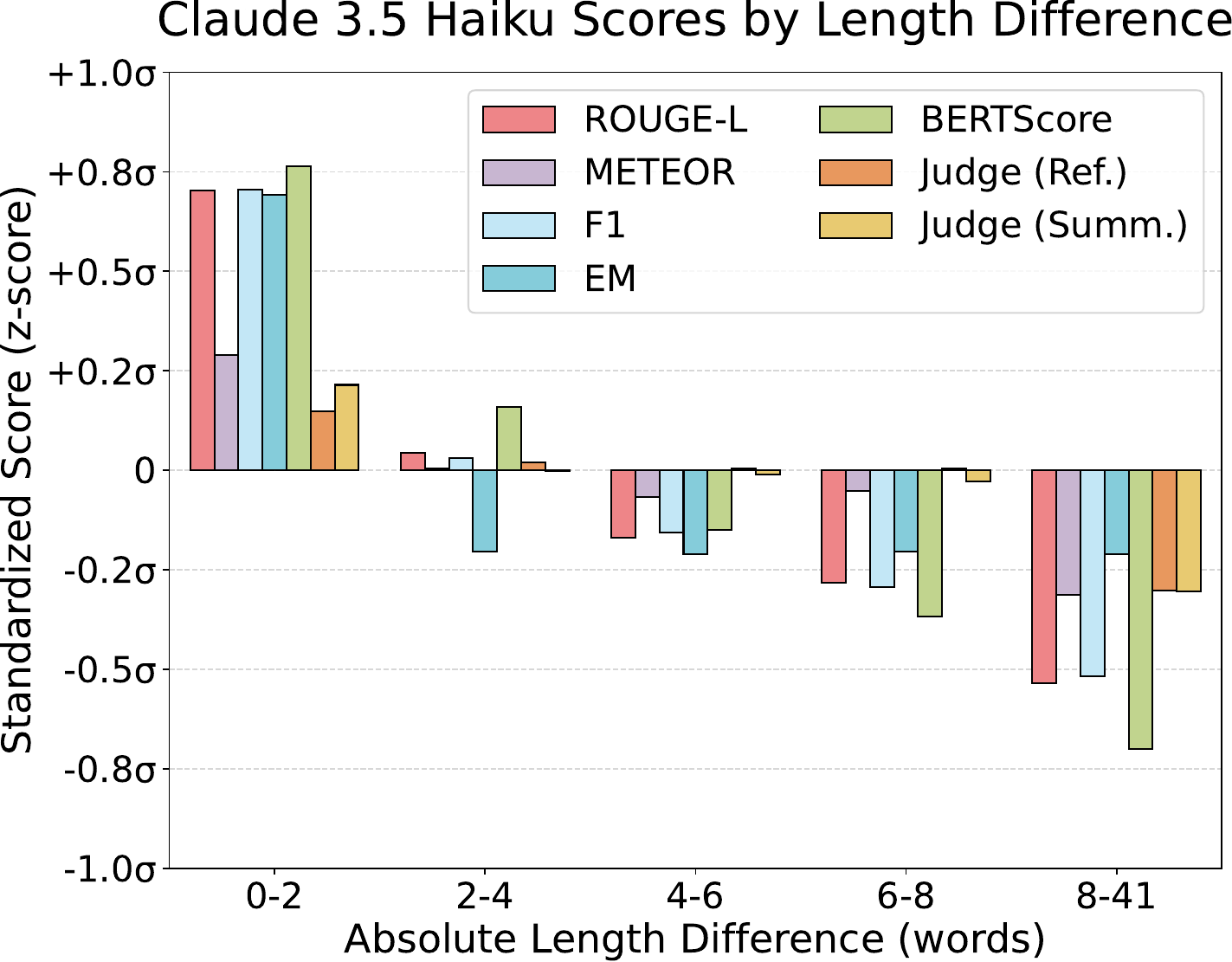}
    \caption{Standardized metrics of Claude 3.5 Haiku's generated responses grouped by the absolute length difference between the prediction and the reference answers. 
    Each bin contains an equal number of samples.}
    \label{fig:grouped_scores_by_pred-ref-diff}
\end{figure}

\paragraph{LLM-as-a-judge}
LLMs used as judges show the highest correlation with human judgments.
When given only the question and reference answers as context to score the predicted answer (reference-based), each LLM achieves a moderately good correlation with human judgments on LiteraryQA, around 0.36 for Sonnet, 0.32 for GPT 4.1 and 0.45 for Prometheus 2.
This indicates that our pipeline has succeeded in improving the quality of QA pairs.
Notably, every LLM shows \textit{consistently higher correlations} on LiteraryQA compared to NarrativeQA in this setting. 
This gap is particularly evident in Prometheus 2, with an increase of 25 percentage points.
It is a stark contrast with NarrativeQA, where we hypothesize that its smaller 7B parameter size may hinder its ability to handle the dataset's noisy reference answers. 
Instead, when provided with LiteraryQA's clean reference answers, it reaches a higher correlation with human judgments than Sonnet 3.7 or GPT 4.1.

When considering the summary-based setting, all system-level correlations increase drastically, arriving at a maximum value of 0.68 for Prometheus 2.
It is clear that letting the judge LLM consider the summary frees it from the restrictions of the reference answers, as the question could accept multiple valid answers within the context of the whole book, as represented by the summary.
In this setting as well, correlations on LiteraryQA are higher than on NarrativeQA, further confirming the effectiveness of our refinement process. 

We conclude that summary-based LLM-as-a-judge has a higher correlation than any of the analyzed \textit{n}-gram- or neural-based metrics on LiteraryQA, and advocate for its use in future work.

Finally, in \Cref{fig:grouped_scores_by_pred-ref-diff} we show how the length difference in words between the generated answers and the references impacts the (standardized\footnote{We standardize a metric's scores by subtracting its the mean and dividing by its the standard deviation.}) metric scores. 
Most metrics show greater score variability at extreme length differences. BERTScore, ROUGE-L, and F1 are particularly sensitive, assigning higher scores for small length mismatches and lower scores for large ones. EM scores remain consistently low due to its binary nature. Among n-gram metrics, only METEOR maintains stable performance across all bins, behaving similarly to Prometheus 2 7B, which has the lowest variability and highest correlation with human judgments.




\begin{table*}[t]
\centering
\begin{tabularx}{\linewidth}{Xccccccc}
\hline
\multirow{2}{*}{Dataset} &
  \multirow{2}{*}{\# Docs} &
  \multicolumn{6}{c}{Claude 3.5 Haiku} \\
  \cline{3-8}
 & & R-1 & R-2 & R-L & METEOR & EM & F1 \\ \hline 
NarrativeQA (Original) & 177 &  0.2208 &  0.0771 &  0.2079 &  0.2743 &  0.0117 &  0.2380 \\
NarrativeQA (Filtered) & 138 &  0.2305 &  0.0824 &  0.2174 &  0.2855 &  0.0122 &  0.2480 \\
LiteraryQA             & 138 &  \textbf{0.2655} &  \textbf{0.1037} &  \textbf{0.2534} &  \textbf{0.2989} &  \textbf{0.0425} &  \textbf{0.2818} \\
\hline
\multirow{2}{*}{Dataset} &
  \multirow{2}{*}{\# Docs} &
  \multicolumn{6}{c}{Gemini 2.0 Flash Lite} \\
  \cline{3-8}
 &   & R-1 & R-2 & R-L & METEOR &  EM &  F1 \\ \hline 
NarrativeQA (Original) &  177 &  0.2307 &  0.0805 &  0.2201 &  0.2635 &  0.0237 &  0.2522 \\
NarrativeQA (Filtered) &  138 &  \textbf{0.2402} &  0.0850 &  0.2294 &  0.2745 &  \textbf{0.0254} &  \textbf{0.2612} \\
LiteraryQA             &  138 &  0.2399 &  \textbf{0.0863} &  \textbf{0.2300} &  \textbf{0.2827} &  0.0158 &  0.2575 \\
  \hline
\end{tabularx}
\caption{Performance increase of closed models responses (Claude 3.5 Haiku and Gemini 2.0 Flash Lite) evaluated through \textit{n}-gram-based metrics across NarrativeQA, NarrativeQA Filtered, and LiteraryQA.}
\label{tab:LQA_vs_NQA_performance}
\end{table*}

\section{LLM benchmarking}
\begin{figure}[t]
\includegraphics[width=\columnwidth]{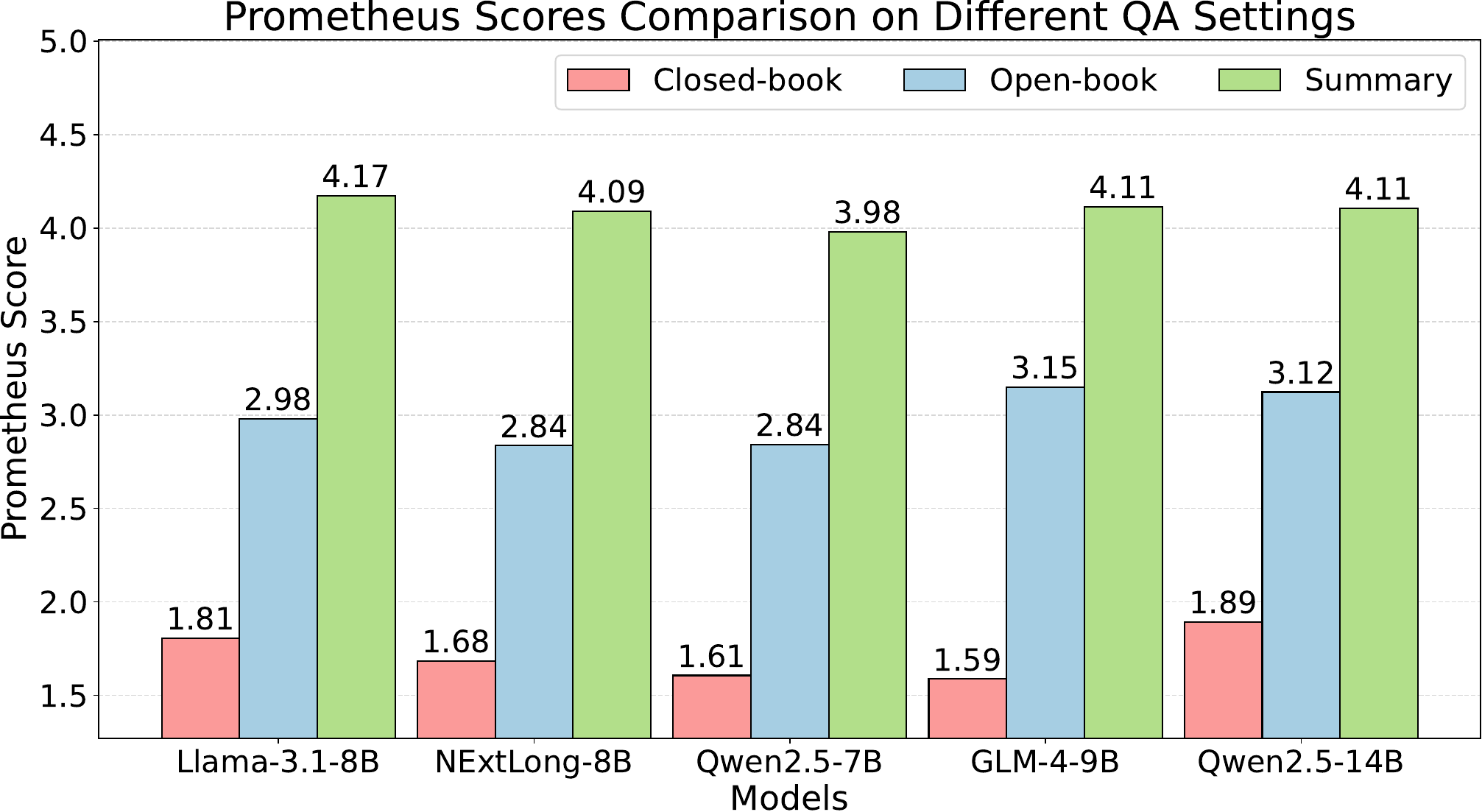}
\caption{Prometheus-as-a-judge scores of the models across different settings.}
\label{fig:Prometheus-performance-baselines}
\end{figure}

In this section, we report the performance on the test set of LiteraryQA of the models in \Cref{tab:models_stats} across three distinct settings. 
In the \textbf{open-book setting}, models have access to the complete narrative text, testing their ability to locate and integrate relevant information across extensive narratives.
We report the performance according to all metrics in the open-book setting in \Cref{tab:lqa-performance}.
Three out of the four \textit{n}-gram-based metrics (ROUGE-L, EM, F1) rank the systems in the same order, with NExtLong-8B achieving higher scores than all other models, including closed-source ones.
Perhaps surprisingly, BERTScore follows the same trend as these metrics.
As described in the previous section, we note that a lower score in \textit{n}-gram-based metrics does not necessarily imply a wrong output, but merely that the generated answer was \textit{syntactically different} from the references.
METEOR, which was identified as the best \textit{n}-gram-based metric, identifies GLM-4-9B as the best model.
The closed-source models are consistently ranked below their counterparts, which contrast with a comparative analysis of a sample of all models' answers.
In fact, according to Prometheus 2 judgments on a sample of the predictions, the best performing model is a closed one, Claude 3.5 Haiku; however, the other closed model, Gemini 2.0 Flash Lite, is not among the top scoring ones (scores reported as ``open-book'' in \Cref{fig:Prometheus-performance-baselines}).

In addition to the evaluation of the performance of the models on LiteraryQA, we establish comparative baselines on both the complete book section of NarrativeQA and the filtered subset containing only the 138 documents included in LiteraryQA.
The results in \Cref{tab:LQA_vs_NQA_performance} show that the predictions of closed-source models become progressively more similar to the reference answers following the steps of our pipeline, as measured by \textit{n}-gram-based metrics.
This suggests a reduction in noise in LiteraryQA compared to NarrativeQA.




We also test the models in two other settings.
In the \textbf{closed-book setting}, models receive only the literary work's title, without additional context, requiring them to rely entirely on their pre-training knowledge and limiting their overall performance.
Instead, in the \textbf{summary setting}, models receive story summaries. 
This is the easiest setting, as summaries are brief (typically <500 words) and many answers appear nearly verbatim. Performance differences across settings, according to Prometheus 2, are shown in \Cref{fig:Prometheus-performance-baselines}.

\section{Conclusions}
In this work, we introduced LiteraryQA, 
a human- and LLM-improved subset of NarrativeQA focused exclusively on literary works, addressing the limitations of existing narrative QA datasets. 
The resulting dataset exhibits improved question clarity, reduced ambiguity, and better alignment between questions and reference answers.
Our extensive benchmarking demonstrates that the higher quality of LiteraryQA enables a more reliable and fair evaluation: tested models achieve higher scores in all metrics, and these metrics better reflect human judgments. 
We then carry out a meta-evaluation of automatic metrics, through which we identify METEOR as the most reliable among \textit{n}-gram approaches, though LLM-as-a-judge systems demonstrated a significantly higher correlation with human judgments when provided with the book summaries.
However, despite these improvements, overall performance remains below that observed in other QA settings, indicating that  LiteraryQA (and in general the open-ended narrative QA setting) continues to represent a challenging benchmark for reading comprehension tasks.



\section*{Limitations}
While LiteraryQA improves the quality and reliability of NarrativeQA, several limitations remain. 

First, the refinement process relies partly on an LLM to support human validation, which can introduce potential biases. 
Although human oversight mitigates this to some extent, the final dataset may still reflect these biases and subjective interpretations of question validity and answer correctness. 

Second, our subset focuses exclusively on literary works, excluding other narrative forms such as movie scripts and theatrical plays. 
While this design choice supports our goal of creating a reliable and homogeneous benchmark, the resulting dataset should not be taken as representative of the \textit{full} narrative landscape.

Third, we did not include any retrieval-augmented generation (RAG) approaches in our evaluations, as our focus was on assessing the ability of the models to comprehend and reason over the entire narrative texts. 
Although RAG methods could potentially enhance performance by retrieving relevant context, they introduce additional complexity and issues that are orthogonal to our goal of evaluating narrative understanding.
Retrieving small fragments can disrupt the narrative flow, which is critical for tasks where coherence and temporal structure are essential.
Exploring RAG in this setting remains an interesting direction for future work. 

Finally, LLM-as-a-judge evaluations, despite showing stronger alignment with human assessments, are i) costly to run at scale, and ii) lack transparency, posing challenges for reproducibility and standardization.
While small fine-tuned models like Prometheus have proven helpful even in this out-of-domain setting, we believe that models specifically fine-tuned for narrative evaluation could offer more accurate and cost-effective alternatives, especially if supported by structured knowledge or grounded in an external knowledge base, enabling more consistent and context-aware judgments.

\section*{Acknowledgments}
\begin{center}
    \noindent
    \begin{minipage}{0.1\linewidth}
        \includegraphics[scale=0.04]{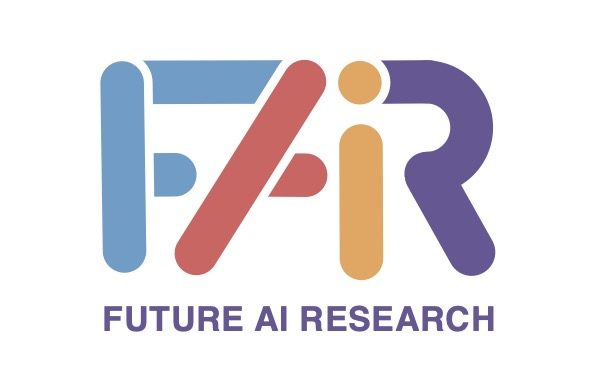}
    \end{minipage}
    \begin{minipage}{0.75\linewidth}
    The authors acknowledge the support of the PNRR MUR project PE0000013-FAIR. 
    \end{minipage}
    \begin{minipage}{0.1\linewidth}
        \includegraphics[scale=0.08]{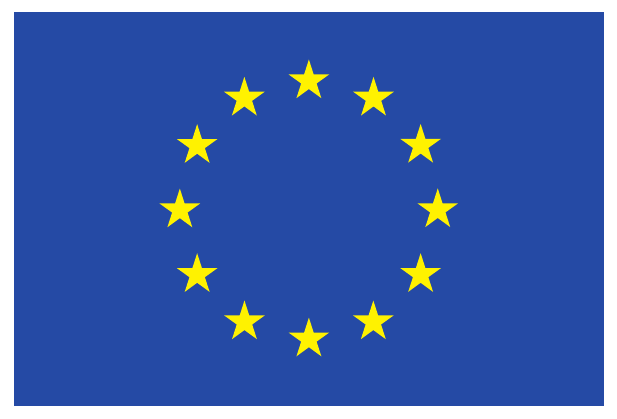}
    \end{minipage}
\end{center}
\vspace{0.2cm}
The authors gratefully acknowledge the support of the AI Factory IT4LIA project and the CINECA award IsCc8\_CRAFT under the ISCRA initiative for granting access to high-performance computing resources.
Finally, the authors would also like to thank Alessandro Scirè for his insights and contributions in the initial phase of the project.

\bibliography{anthology-1,anthology-2,custom}

\appendix

\section{Comparison with NovelQA}\label{app:novelqa}
To better contextualize LiteraryQA's contributions, we compare it with NovelQA~\citep{wang2025novelqa}, a contemporary work that introduces a benchmark for long-form reading comprehension. 
While both datasets target literary understanding, they differ significantly in their design philosophy and evaluation focus (\Cref{tab:literaryqa_novelqa_comparison}).

\begin{table}[h!]
\centering
\begin{tabularx}{\linewidth}{Xcc}
\hline
Aspect & LiteraryQA & NovelQA \\
\hline
Length ($\mu\pm\sigma$)   & 74K $\pm$ 59K & 179K $\pm$ 145K\\
\# Documents  & 138 (test set) & 89\\
QA source & Summaries & Paragraphs \\
Availability & Full & Partial\\
\hline
\end{tabularx}
\caption{Breakdown of the main differences between LiteraryQA and NovelQA. We take into account only the test set of LiteraryQA and the public domain books of NovelQA for the length stats in this Table.}
\label{tab:literaryqa_novelqa_comparison}
\end{table}

NovelQA features documents with an average length approximately 2.5 times longer than those in LiteraryQA. 
However, LiteraryQA contains 50\% more documents than NovelQA, providing broader coverage across different literary works. 
Moreover, the datasets employ fundamentally different annotation approaches. 
NovelQA generates QA pairs through a combination of templates and free-form generation, allowing for systematic coverage but potentially limiting question diversity and introducing biases. 
In contrast, LiteraryQA adopts a fully manual QA generation process (inherited from NarrativeQA) followed by an automatic QA pairs refinement (detailed in \Cref{sec:literaryqa}), which is fundamental in order to obtain more nuanced and varied questions.

Most significantly, the datasets test different aspects of reading comprehension. 
NovelQA focuses on detailed questions supported by specific annotated evidence paragraphs, emphasizing precise information extraction and evidence-based reasoning. 
On the other hand, LiteraryQA derives its questions from book summaries, requiring models to i) synthesize information across multiple passages, ii) avoid full reliance on surface-level pattern matching, and iii) demonstrate broader narrative understanding. 
This distinction makes LiteraryQA particularly suited for evaluating synthesis and reasoning capabilities rather than fine-grained information extraction.

Finally, NovelQA's answers and evidence paragraphs are not publicly released to prevent data contamination in model training, an increasingly legitimate concern for keeping the benchmark integrity.
Moreover, approximately one-fourth of the dataset is composed of copyrighted books, which are not freely accessible. 
However, this design choice also limits researchers' ability to conduct experiments, detailed error analysis, and iteration on evaluation methods. 
In contrast, LiteraryQA, following the open-access approach of its parent dataset NarrativeQA, prioritizes transparency and reproducibility, enabling more comprehensive model analysis and community engagement.

These complementary approaches suggest that both datasets serve important but distinct roles in evaluating literary reading comprehension, with LiteraryQA particularly well-suited for assessing synthesis and narrative understanding capabilities.

\section{Text Extraction Algorithm Details \label{app:pipeline_details}}
\Cref{algo:html_cleaning} shows how we parse the raw HTML Gutenberg data into clean text documents.
We only committed changes to this extraction algorithm when they maintained complete preservation of all narrative samples. 
This conservative approach guaranteed that no valuable narrative content was inadvertently removed.

The algorithm takes as input an HTML document $H$ and a set of parameters $\theta$. 
The $\theta$ parameters contain lists and mappings of HTML tags categorized by their processing needs, such as tags to decompose, unwrap, replace, or remove attributes from.
We use the Python library BeautifulSoup\footnote{https://beautiful-soup-4.readthedocs.io} for parsing the HTML into a tree structure $S$.
Each tag $t$ in the tree $S$ is processed based on its category. 
We define the categories for tags to keep or remove after comparing the source HTML and the rendered page, defining the following cases:
\begin{itemize}
    \item If $t$ is in the Decompose list, it is removed entirely from $S$.
    \item If $t$ is in the Unwrap list, it is replaced by its children nodes in $S$, effectively removing the tag but preserving its content.
    \item If $t$ is in the Replace mapping, the tag $t$ is replaced by the corresponding substitute defined in $\theta$ (e.g., to keep it with a specific formatting).
    \item If $t$ is in the Remove attributes list, specified attributes are removed from $t$ without deleting the tag itself.
\end{itemize}
After modifying the HTML structure in this way, the algorithm extracts text content only from the modified tree structure $S$ and normalizes the text by removing multiple spaces and line breaks and filtering out empty or invalid strings.

T is then processed by \Cref{algo:text_cleaning}, which filters remaining noise that passed the initial step. 
The algorithm processes each line using RegEx patterns to determine whether to skip the line, keep it, reinitialize the clean text buffer, or terminate processing.
Since terminating sequences are more difficult to handle reliably than starting patterns, we introduce a parameter $\alpha$
to control when the algorithm may terminate. 
Specifically, when an ending pattern (e.g., \verb|"THE END."|) is detected, the algorithm stops only if it occurs in the final portion of the text; otherwise, it retains the line and continues processing. 
We set $\alpha=0.9$ in our experiments, allowing termination only when such patterns appear in the last 10\% of the text\footnote{Our implementation also allows certain specific, unambiguous patterns to terminate the process regardless of the $\alpha$
threshold.}.

\begin{algorithm}[t]
\caption{Extract structured text from HTML.}
\label{algo:html_cleaning}
\KwIn{HTML document $H$, options $\theta$}
\KwOut{Raw text $T$}

$S \gets $ parse($H$)\;

\For{tag $t \in S$}{
    \If{$t \in \theta_{\text{decompose}}$}{
        $S \gets S \setminus \{t\}$\;
    }
    \If{$t \in \theta_{\text{unwrap}}$}{
        $S \gets S \cup \text{children}(t)$\;
        $S \gets S \setminus \{t\}$\;
    }
    \If{$t \in \theta_{\text{replace}}$}{
        $t \gets \theta_{\text{replace}}(t)$\;
    }
    \If{$t \in \theta_{\text{remove\_attrs}}$}{
        $t \gets t \setminus \theta_{\text{remove\_attrs}}(t)$\;
    }
}
$T \gets \emptyset$\;
\For{tag $t \in S$}{
    $s \gets \text{normalize(}t_{text}\text{)}$\;
    \If{$s \neq \emptyset \wedge valid$}{
        $T \gets T \cup \{s\}$\;
    }
}
\Return $T$
\end{algorithm}

\begin{algorithm}[t]
\caption{Clean extracted text with start, end, and skip patterns.}
\label{algo:text_cleaning}
\KwIn{Raw text $T$, patterns $\Phi$, threshold $\alpha$}
\KwOut{Cleaned text lines $C$}

$C \gets$ [ ]\;
$L \gets$ split(T)\;

\For{line $l_i \in L$}{
    \If{$l_i \in \Phi_{start}$}{
        $C$ $\gets$ [ ]\;
    }
    \ElseIf{$l_i \in \Phi_{skip}$}{
        continue\;
    }
    \ElseIf{$l_i \in \Phi_{end} \wedge i >= \alpha\cdot |L|$}{
        break\;
    }
    \Else{
        $C \gets C \cup \{l_i\}$\;
    }
}
\Return $C$
\end{algorithm}

\begin{table*}
    \centering
    \begin{tabularx}{\linewidth}{|X|}
        \hline
        \textbf{Question Criteria} \\
        \hline
        An acceptable question must: \\
        \begin{enumerate}
            \item Be grammatically correct in relation to the summary
            \item Be unambiguous and have a clear answer within the summary context
            \item Be answerable using only information present in the summary
        \end{enumerate}\\
        \hline
        \textbf{Answer Criteria}\\
        \hline
        An acceptable answer must:\\
        \begin{enumerate}
        \item Be grammatically correct, specifically: \begin{enumerate}
            \item Free of typos
            \item Free of misspellings
            \item Free of mistakes due to accidental key presses
            \item Include proper contractions and possessives (e.g., \textit{don't}, \textit{wasn't}, \textit{John's})
            \item Use correct subject-verb agreement
        \end{enumerate}
        \item Be factually correct and complete according to the summary:
        \begin{enumerate}
            \item Contain no information contradicting the summary
            \item Include all relevant entities (people, locations, dates, etc.) when applicable
            \item Provide a single, precise response (not multiple possibilities or vague statements)
        \end{enumerate}
        \item Be properly scoped:
        \begin{enumerate}
            \item Include only information found in the summary
            \item Be concise while addressing the full question
            \item Avoid speculation beyond what's stated in the summary
        \end{enumerate}
        \end{enumerate}\\
        \hline
    \end{tabularx}
    \caption{Guidelines used throughout the annotation process.}
    \label{tab:annotation_guidelines}
\end{table*}

\section{Additional Results on LiteraryQA}\label{app:LQA_stats}
\begin{figure}[t]
\includegraphics[width=\columnwidth]{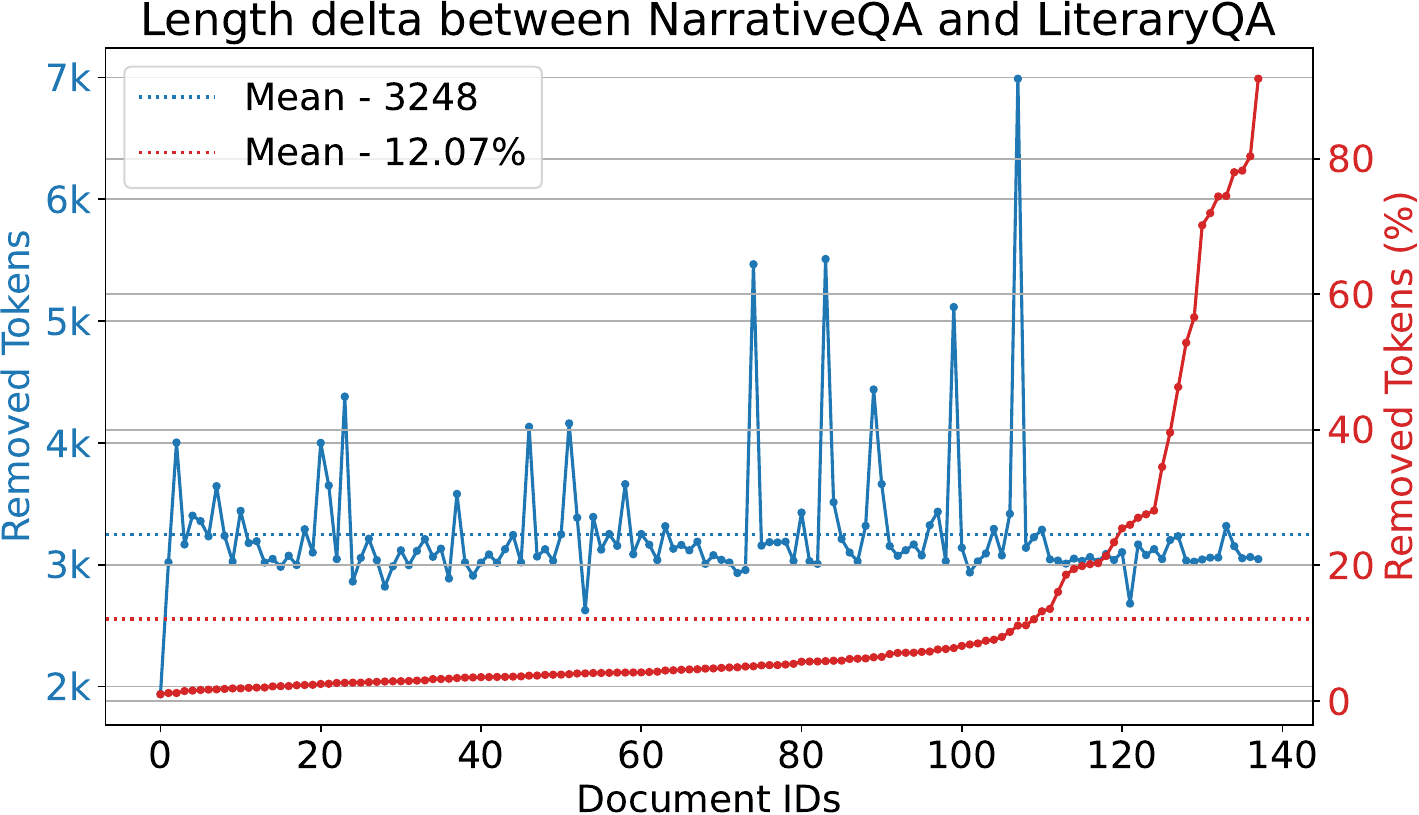}
\caption{Comparison of document lengths between NarrativeQA and LiteraryQA for the same books. Blue line indicates absolute token differences; red line shows percentage differences.}
\label{fig:token_count_LQAvsNQA}
\end{figure}

In this Section we present additional details on final version of the dataset, LiteraryQA.
\Cref{fig:token_count_LQAvsNQA} shows the length difference in tokens between the 138 shared documents in LiteraryQA and NarrativeQA.
Our cleaning procedure removes an average of 3K tokens per document, representing 12\% of the original text.

\Cref{tab:doc_pipeline_performance} presents the classification performance of our pipeline's document-level steps using Llama-3.1-8B-Instruct on the test set.
The clear-cut nature of this classification task enables perfect agreement between evaluators.

\begin{table}[t]
\centering
\begin{tabularx}{\linewidth}{Xccccc}
\hline
Category & Pr & Re & F1 & $\kappa$ \\
\hline
Mismatched     & 0.99 & 0.73 & 0.81 & 1.00 \\
Plays          & 1.00 & 1.00 & 1.00 & 1.00 \\
Non-narrative  & 0.86 & 0.79 & 0.82 & 1.00 \\
\hline
\end{tabularx}
\caption{Classification performance of Llama-3.1-8B-Instruct on the documents categorized by the annotator showing Precision (Pr), Recall (Re) and F1-score (F1). We also report the Inter-Annotator Agreement through Cohen's Kappa.}
\label{tab:doc_pipeline_performance}
\end{table}

The complete list of the 20 annotated documents (583 QA samples) can be found in \Cref{tab:annotated_docs}.
We chose these books because they span over multiple genres, authors, styles, and languages (a few were originally written in French, although we only work on the English versions). 

\begin{table*}[h!]
\centering
\begin{tabularx}{\linewidth}{rXXl}
 & \textbf{Title} & \textbf{Author} & \textbf{Nationality} \\
\hline
1.& \textit{A Portrait of the Artist as a Young Man} & James Joyce & Irish \\
2.& \textit{A Voyage to Arcturus} & David Lindsay & Scottish \\
3.& \textit{Father Goriot} & Honoré de Balzac & French \\
4.& \textit{Lisbeth Longfrock} & Hans Aanrud & Norwegian \\
5.& \textit{Lothair} & Benjamin Disraeli & British \\
6.& \textit{Peter Pan in Kensington Gardens} & J. M. Barrie & Scottish \\
7.& \textit{Tarzan and the Jewels of Opar} & Edgar Rice Burroughs & American \\
8.& \textit{Tarzan of the Apes} & Edgar Rice Burroughs & American \\
9.& \textit{The Adventures of the Dying Detective} & Arthur Conan Doyle & Scottish \\
10.& \textit{The Black Dwarf} & Walter Scott & Scottish \\
11.& \textit{The Call of the Wild} & Jack London & American \\
12.& \textit{The Children of the New Forest} & Frederick Marryat & British \\
13.& \textit{The House of the Seven Gables} & Nathaniel Hawthorne & American \\
14.& \textit{The House on the Borderland} & W. H. Hodgson & British \\
15.& \textit{The Gods Are Athirst} & Anatole France & French \\
16.& \textit{The Vampyre} & John Polidori & British \\
17.& \textit{The Variable Man} & Philip K. Dick & American \\
18.& \textit{Uncle Silas} & Joseph S. Le Fanu & Irish \\
19.& \textit{Voodoo Planet} & Andre Norton & American \\
20.& \textit{Youth} & Joseph Conrad & Polish-British \\
\hline
\end{tabularx}
\caption{Subset of annotated documents for the evaluation of the data refinement pipeline.}
\label{tab:annotated_docs}
\end{table*}

\begin{figure}[t]
    \includegraphics[width=\columnwidth]{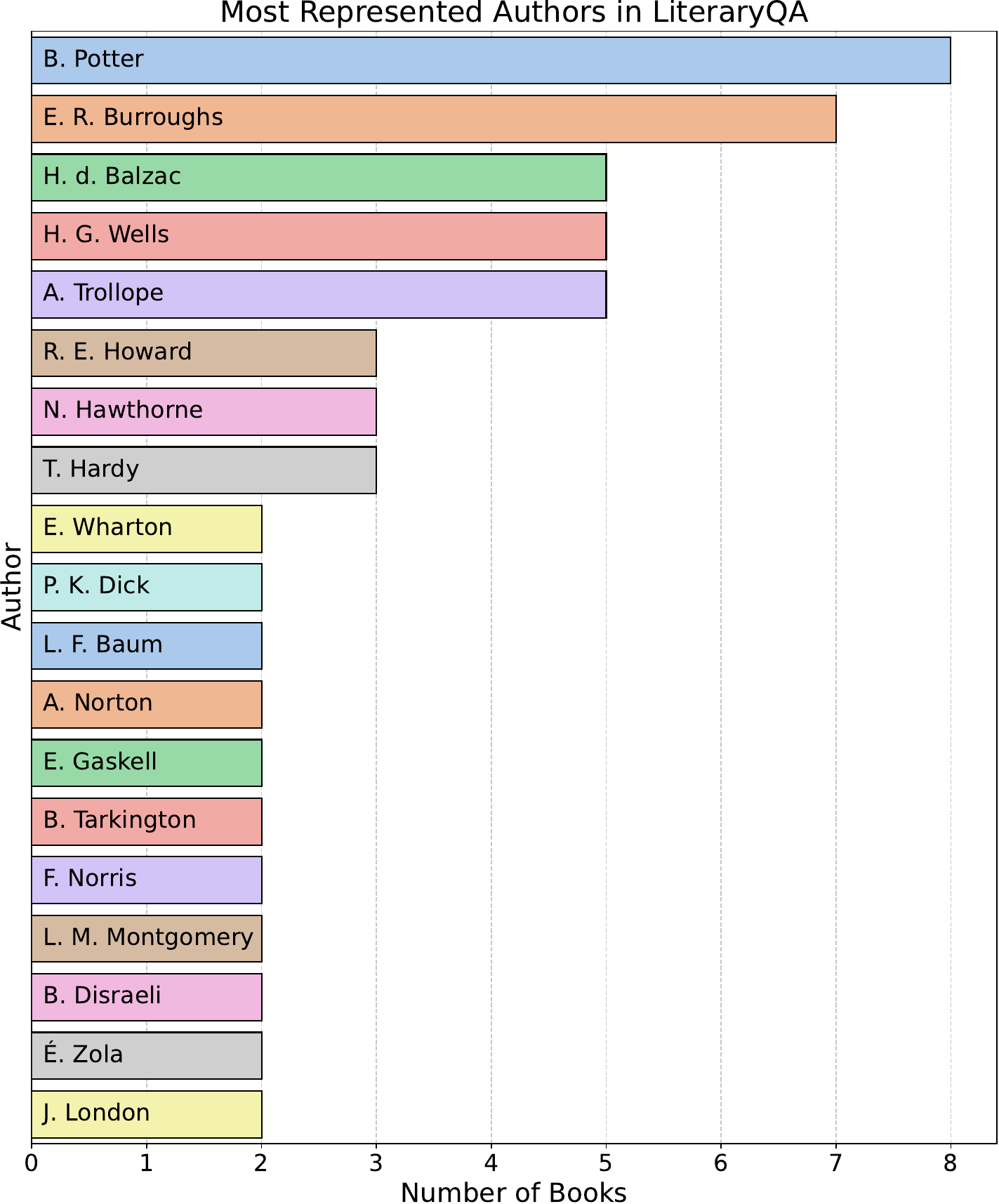}
    \caption{List of the most represented authors in LiteraryQA (test set).}
    \label{fig:top_authors}
\end{figure}

\begin{figure}[t]
    \includegraphics[width=\columnwidth]{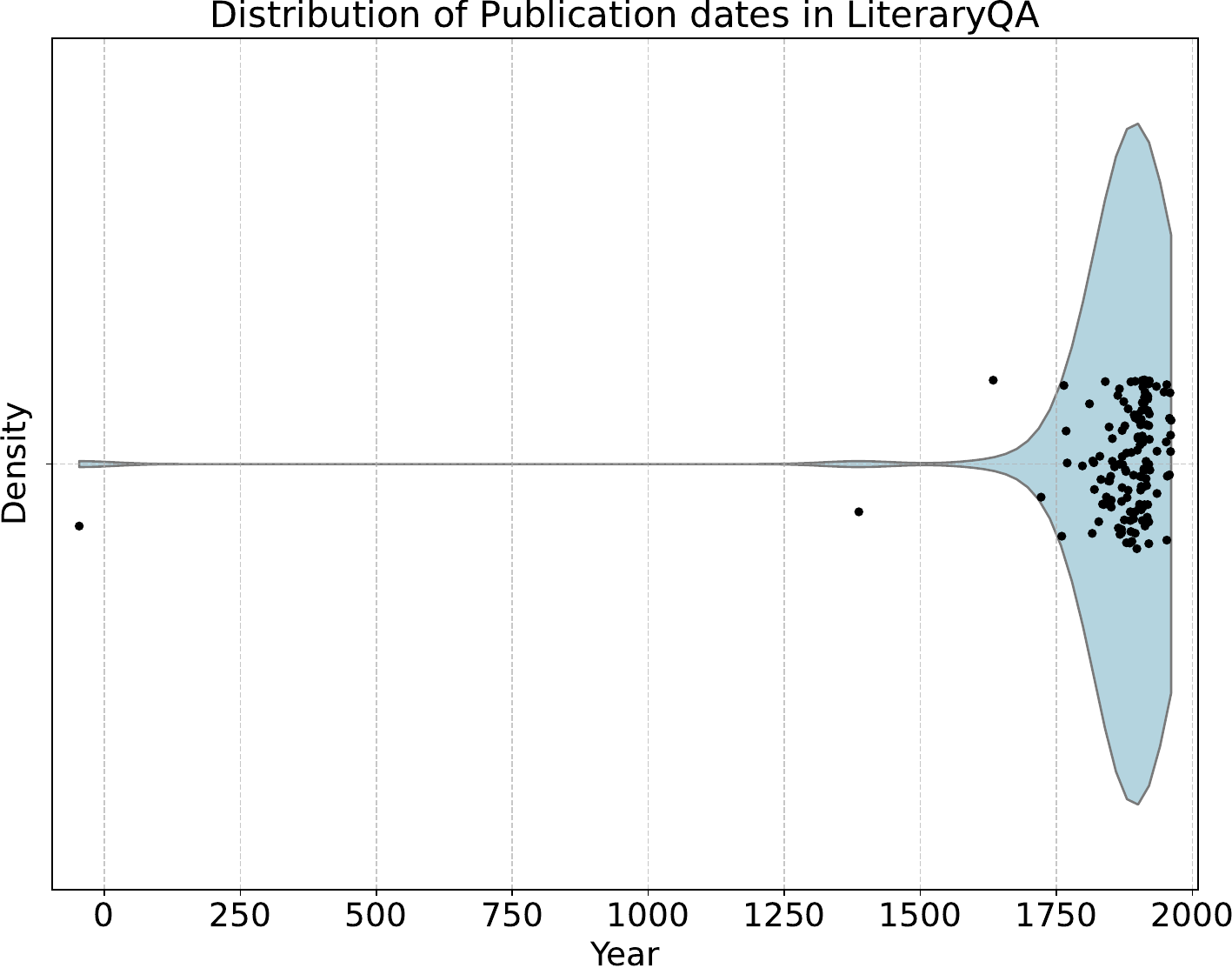}
    \caption{Publication year distribution in LiteraryQA (test set).}
    \label{fig:year_distribution_lqa}
\end{figure}

\Cref{fig:top_authors,fig:year_distribution_lqa} show the most represented authors and the publication years within the LiteraryQA test set, respectively.

Finally, we report the prompt we used throughout the data refinement pipeline in \Cref{tab:prompt-questions,tab:prompt-answers,tab:categorizer_prompt_play,tab:categorizer_prompt_mismatched,tab:categorizer_prompt_non_fiction}, and some examples of the corrections made by Claude 3.5 Haiku in \Cref{tab:claude_corrections}.

\begin{table*}[!t]
\centering
\begin{tabularx}{\linewidth}{|X|X|}
\hline
\textbf{Original Sample} & \textbf{Refined Sample}\\
\hline

\multicolumn{2}{|>{\hsize=\dimexpr2\hsize+2\tabcolsep+\arrayrulewidth\relax}X|}{\textbf{Summary:} [...] The rats abandon the reshipped barque and \colorbox{nicegreen}{a new crew is brought in from Liverpool}}\\
& \\
\textbf{Q:} Why is a new \colorbox{nicered}{ship} brought in from Liverpool? & \textbf{Q:} Why was a new \colorbox{nicegreen}{crew} brought in from Liverpool? \\
\textbf{A1:} because no man will stay on a ship abandoned by rats & \textbf{A1:} because no man will stay on a ship abandoned by rats\\
\textbf{A2:} The ship had been abandoned by rats. & \textbf{A2:} The ship had been abandoned by rats. \\
\hline

\multicolumn{2}{|>{\hsize=\dimexpr2\hsize+2\tabcolsep+\arrayrulewidth\relax}X|}{\textbf{Summary:} [...] Indefer Jones' niece, Isabel Brodrick, has lived with him for years after the remarriage of her father, and endeared herself to everyone. However, according to his strong traditional beliefs, the estate should be bequeathed to a male heir. His sole male blood relative is his nephew Henry Jones [...]}\\
& \\
\textbf{Q:} Why did Indefer Jones originally leave his estate to his nephew when he wanted to leave it to his niece? &
\textbf{Q:} Why did Indefer Jones \colorbox{niceblue}{initially plan} to leave his estate to \colorbox{niceblue}{ his male heir} \colorbox{niceblue}{Henry despite preferring his niece Isabel}?\\
\textbf{A1:} Because that was tradition &
\textbf{A1:} Because that was tradition \\
\textbf{A2:} Because that was tradition &
\textbf{A2:} Because that was tradition \\
\hline

\multicolumn{2}{|>{\hsize=\dimexpr2\hsize+2\tabcolsep+\arrayrulewidth\relax}X|}{\textbf{Summary:} [...] While Oxford's academic staff barely notice that nearly all of their undergraduates have vanished, \colorbox{nicegreen}{Zuleika decides to order a special train for the next morning} [...]}\\
& \\
\textbf{Q:} After many of the students have died how does Zuleika choose to travel away from Oxford? & 
\textbf{Q:} After many of the students have died how does Zuleika choose to travel away from Oxford?\\
\textbf{A1:} On a train & \textbf{A1:} On a train\\
\textbf{A2:} \colorbox{nicered}{All of the undergraduate students.} &
\textbf{A2:} \colorbox{nicegreen}{She orders a special train to Cambridge.}\\
\hline

\multicolumn{2}{|>{\hsize=\dimexpr2\hsize+2\tabcolsep+\arrayrulewidth\relax}X|}{\textbf{Summary:} [...] The book begins with the death of Helen Carey, the much beloved mother of nine-year-old Philip Carey [...] he is sent to live with his aunt Louisa and uncle William Carey [...]}\\
& \\
\textbf{Q:} Who was Phillip sent to live with after his mother died? &
\textbf{Q:} Who was Phillip sent to live with after his mother died? \\
\textbf{A1:} His aunt and uncle &
\textbf{A1:} His aunt and uncle\\
\textbf{A2:} Aunt and uncle & 
\textbf{A2:} \colorbox{niceblue}{Philip was sent to live with his aunt Louisa} \colorbox{niceblue}{and uncle William Carey} \\
\hline
\end{tabularx}
\caption{Examples of QA samples from different books after being corrected by Claude.
In the first two samples, Claude corrected the question, while in the last two one of the reference answers.
Factual and semantic errors are highlighted in red, while their correction and evidence is in green. 
Although the second and fourth original samples contained no major errors, Claude improved their grammatical fluency and specificity (highlighted in blue).}
\label{tab:claude_corrections}
\end{table*}

\begin{table*}[t]
\centering
\begin{tabularx}{\linewidth}{|X|}
\hline
\textbf{System Prompt} 
\\
\texttt{Your task is to determine whether a question is not acceptable (grammatically malformed and/or ill-posed with respect to the reference summary).  
The question may refer to unusual, made-up, or technical words found in the reference summary — this is acceptable only if they are spelled consistently.}\\
\texttt{A question is malformed if it contains *any* common grammatical or misspellings errors, for example (non-exhaustive list):}\\
\texttt{- Misspelled words (including names and summary terms spelled inconsistently)}\\
\texttt{- Redundant or conflicting auxiliary verbs (e.g., 'was can not')}\\
\texttt{- Incorrect verb tense or verb form after auxiliaries (e.g., 'did played', 'does belives')}\\
\texttt{- Subject-verb disagreement (e.g., 'whose runs')}\\
\texttt{- Fat-finger errors (e.g., too many or missing whitespaces, letters inversions)}\\
\texttt{- Include proper contractions and possessives (e.g., 'who's', 'it's', 'he's')}\\
\texttt{- Faulty structure (e.g., missing auxiliaries, incorrect use of question words)}\\
\texttt{A question is ill-posed if (non-exhaustive list):}\\
\texttt{- It refers to something (an event, a character, etc.) that is not present in the summary}\\
\texttt{- It misunderstands the summary or misrepresents its content}\\
\texttt{- It does not have a clear answer in the summary}\\
\texttt{A question is well-posed if it is clear, unambiguous, and has a specific answer in the summary.}\\
\texttt{If the question is not acceptable, rewrite it so to keep it as close as possible to the original question, while making it well-formed and well-posed. 
Respond in JSON format with exactly this structure:}\\
\texttt{\{
  "label": "acceptable" or "not acceptable",
  "correction": "..." // rewrite the question with the least amount of edits if it is not acceptable, otherwise write an empty string
\}}\\
\texttt{{Only output this JSON. Do not add any commentary, do not explain your changes.}}\\  
\hline
\textbf{User Prompt}
\\
\texttt{Reference summary: \{summary\}}\\
\texttt{Question: \{question\}}\\
\texttt{Is the question acceptable or not? Follow the rules above and respond with a JSON object as specified.}\\
\hline
\end{tabularx}
\caption{System prompt used with Claude 3.5 Haiku to identify and correct invalid question samples.}
\label{tab:prompt-questions}
\end{table*}

\begin{table*}[t]
\centering
\begin{tabularx}{\linewidth}{|X|}
\hline
\textbf{System Prompt} 
\\
\texttt{You are an English teacher evaluating answers about a narrative.} \\
\texttt{Your task is to determine whether an answer is acceptable (grammatically well-formed and valid).} \\

\texttt{The answer may refer to unusual, made-up, or technical words found in the reference summary — this is acceptable only if they are spelled consistently.} \\
\texttt{An answer is malformed if it contains *any* common grammatical or misspellings errors, for example (non-exhaustive list):} \\
\texttt{- Misspelled words (including names and summary terms spelled inconsistently)} \\
\texttt{- Redundant or conflicting auxiliary verbs (e.g., 'was can not')} \\
\texttt{- Incorrect verb tense or verb form after auxiliaries (e.g., 'did played', 'does belives')} \\
\texttt{- Fat-finger errors (e.g., too many or missing whitespaces, letters inversions)} \\
\texttt{- Include proper contractions and possessives (e.g., 'who's', 'it's', 'he's')} \\
\texttt{- Faulty structure (e.g., missing auxiliaries, incorrect use of question words)} \\

\texttt{A question is valid according to the following criteria:} \\
\texttt{- The answer must be factually correct, i.e. it must be supported by the reference summary, AND} \\
\texttt{- The answer must be complete (include all necessary entities for a complete response), AND} \\
\texttt{- The answer must provide a single precise response, not multiple possibilities or vague statements, AND} \\
\texttt{- The answer must be properly scoped, i.e. it must concisely address the question using the information found in the summary and without speculating or adding information.} \\
\texttt{Finally, the answer may consist of only one or two words — this is acceptable provided that there are no grammatical errors and the above criteria are met.} \\

\texttt{Respond in JSON format with exactly this structure:} \\
\texttt{\{ } \\
\texttt{  "label": "acceptable" or "not acceptable",} \\
\texttt{  "correction": "..." // if "not acceptable", rewrite the answer with the smallest amount of edits to make it acceptable, otherwise write an empty string} \\
\texttt{\}} \\
\texttt{Only output this JSON. Do not add any commentary, do not explain your changes.} \\
\textbf{User Prompt}
\\
\texttt{Reference summary: \{summary\}}\\
\texttt{Question: \{question\}}\\
\texttt{Answer: \{answer\}}\\
\texttt{Is the answer acceptable or not? Follow the rules above and respond with a JSON object as specified.}\\
\hline
\end{tabularx}
\caption{System prompt used with Claude 3.5 Haiku to identify and correct invalid answers samples.}
\label{tab:prompt-answers}
\end{table*}

\begin{table*}[t]
\centering
\begin{tabularx}{\linewidth}{|X|}
\hline
\textbf{System Prompt} 
\\
\texttt{You are an expert literature analyst. }
\texttt{Given a book description, you extract its category (novel or play).}
\texttt{You rely ONLY on the text provided and do not make up any information.}
\\
\textbf{User Prompt}
\texttt{Description: \{description\}}
\texttt{Is this a novel or a play? Reply with one word and do not include any other information.}\\
\hline
\end{tabularx}
\caption{System prompt used with Llama-3.1-8B-Instruct to identify theatrical plays.}
\label{tab:categorizer_prompt_play}
\end{table*}

\begin{table*}[t]
\centering
\begin{tabularx}{\linewidth}{|X|}
\hline
\textbf{System Prompt} 
\\
\texttt{You are an expert literature analyst. }
\texttt{Given a book description, you extract its category (novel or non-fiction).}
\texttt{You rely ONLY on the text provided and do not make up any information.}
\\
\textbf{User Prompt}
\texttt{Description: \{description\}}
\texttt{Is this a novel or a non-fiction? Reply with one word and do not include any other information.}\\
\hline
\end{tabularx}
\caption{System prompt used with Llama-3.1-8B-Instruct to identify non-fiction books.}
\label{tab:categorizer_prompt_non_fiction}
\end{table*}

\begin{table*}[t]
\centering
\begin{tabularx}{\linewidth}{|X|}
\hline
\textbf{System Prompt} 
\\
\texttt{You are an expert literature analyst. }
\texttt{Given a book summary and its first paragraphs, you identify whether the two refer to the same literary work.}
\texttt{You rely ONLY on the text provided and do not make up any information.}
\\
\textbf{User Prompt}
\texttt{Summary: \{summary\}}
\texttt{Paragraphs: \{paragraphs\}}
\texttt{Do they refer to the same literary work? Reply with yes/no and do not include any other information.}\\
\hline
\end{tabularx}
\caption{System prompt used with Llama-3.1-8B-Instruct to identify mismatched samples.}
\label{tab:categorizer_prompt_mismatched}
\end{table*}

\begin{figure}[h]
\includegraphics[width=\columnwidth]{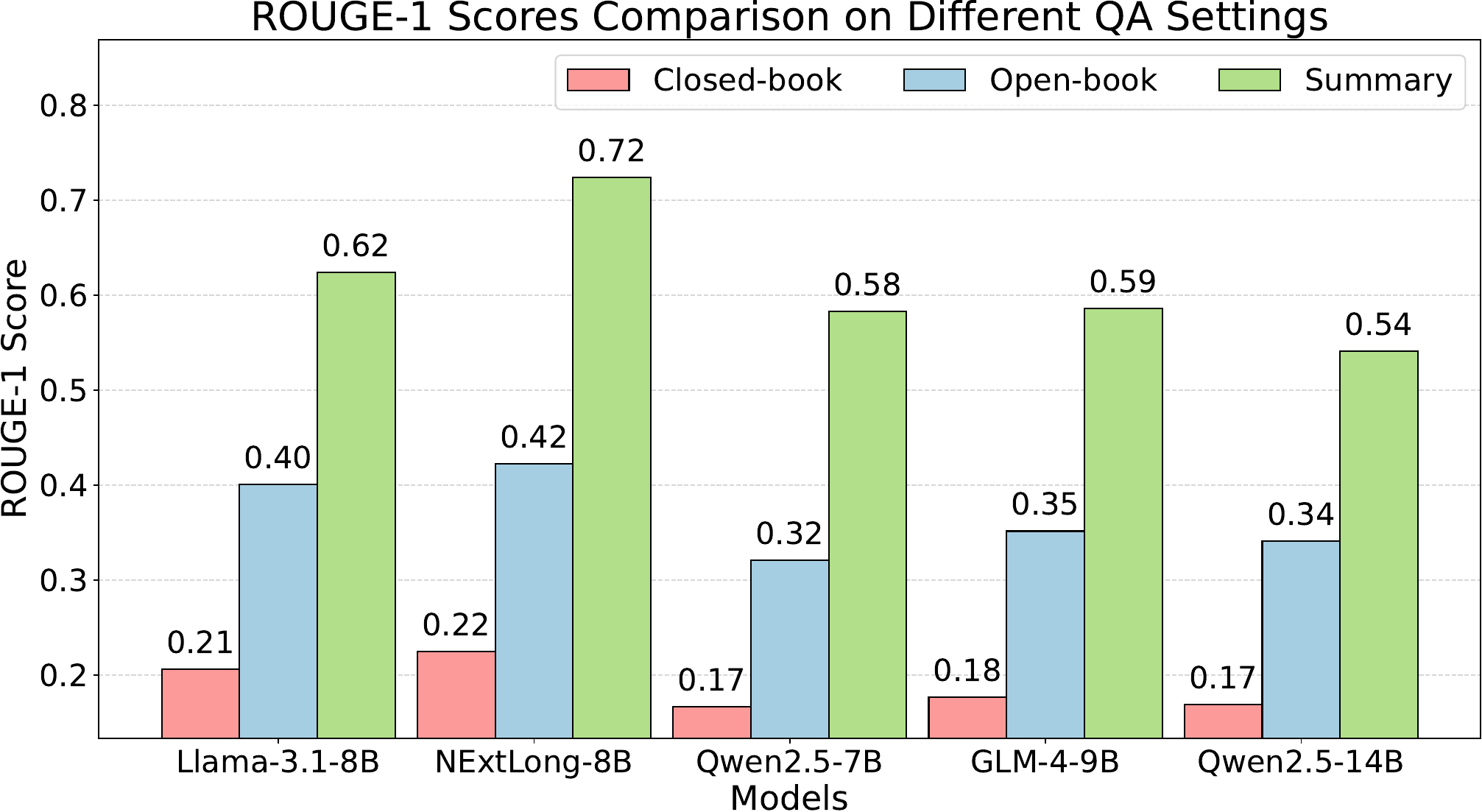}
\caption{ROUGE-1 scores of the models across different settings.}
\label{fig:r1_performance_baselines}
\end{figure}

\begin{figure}[h]
\includegraphics[width=\columnwidth]{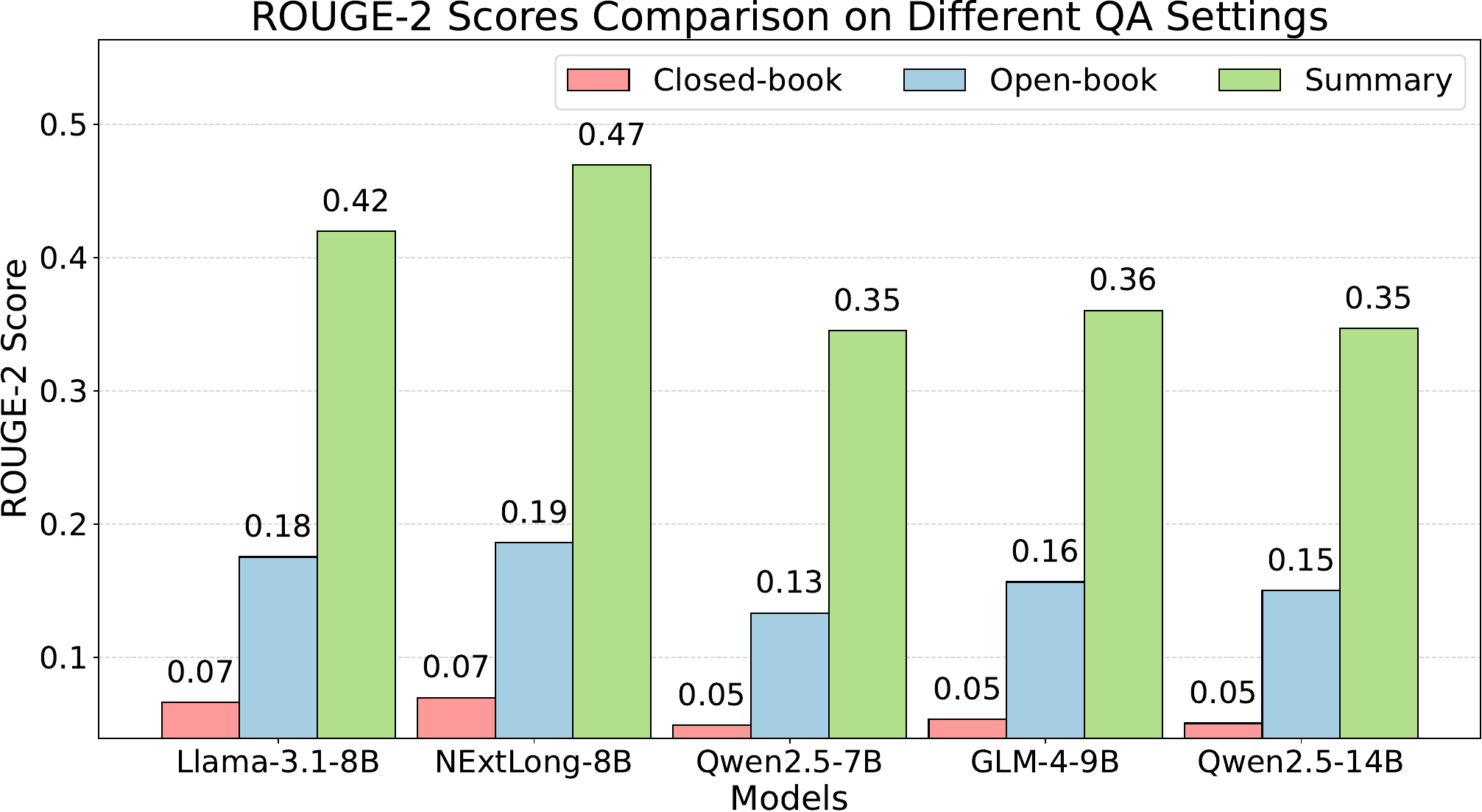}
\caption{ROUGE-2 scores of the models across different settings.}
\label{fig:r2_performance_baselines}
\end{figure}

\begin{figure}[h]
\includegraphics[width=\columnwidth]{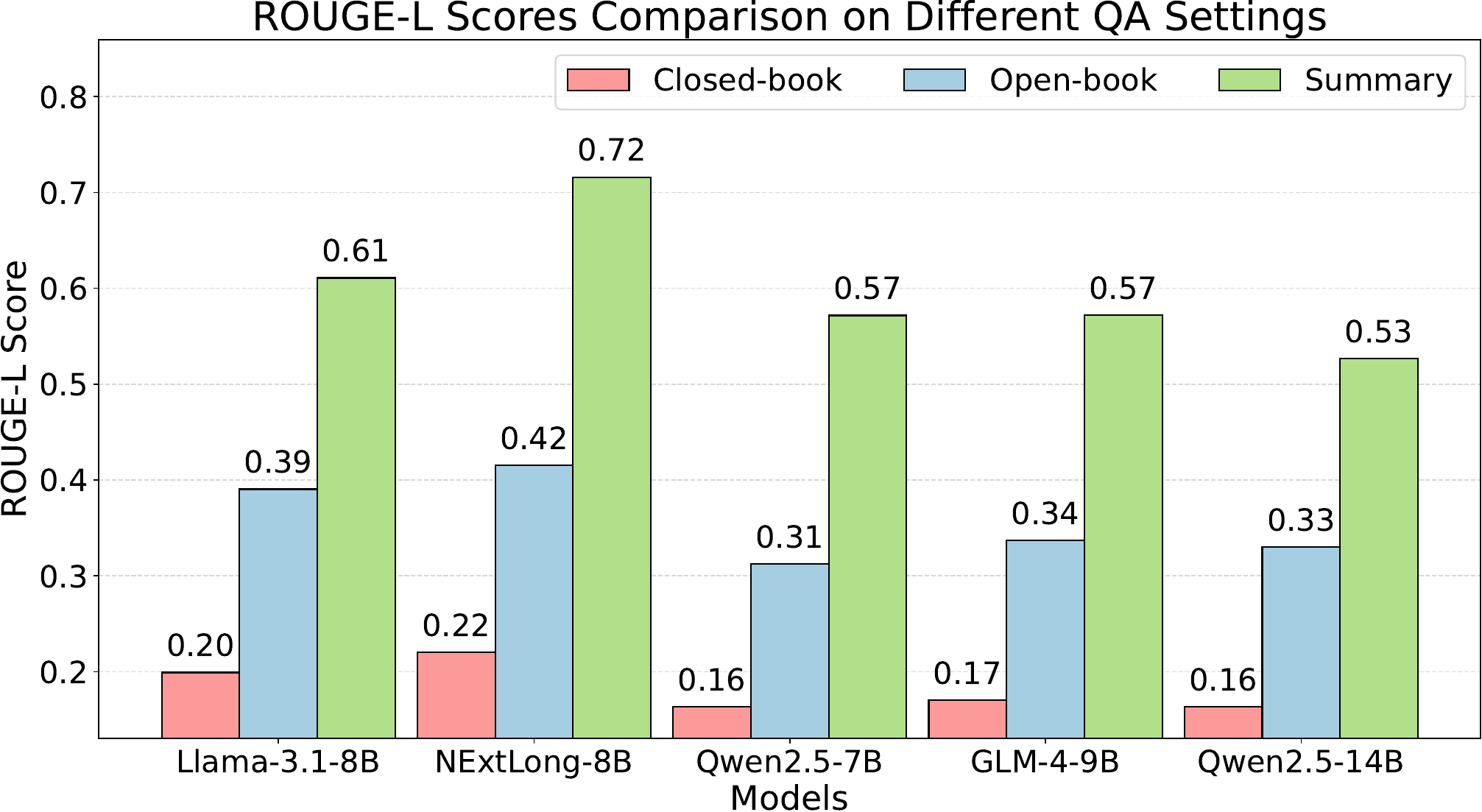}
\caption{ROUGE-L scores of the models across different settings.}
\label{fig:rL-performance_baselines}
\end{figure}

\begin{figure}[h]
\includegraphics[width=\columnwidth]{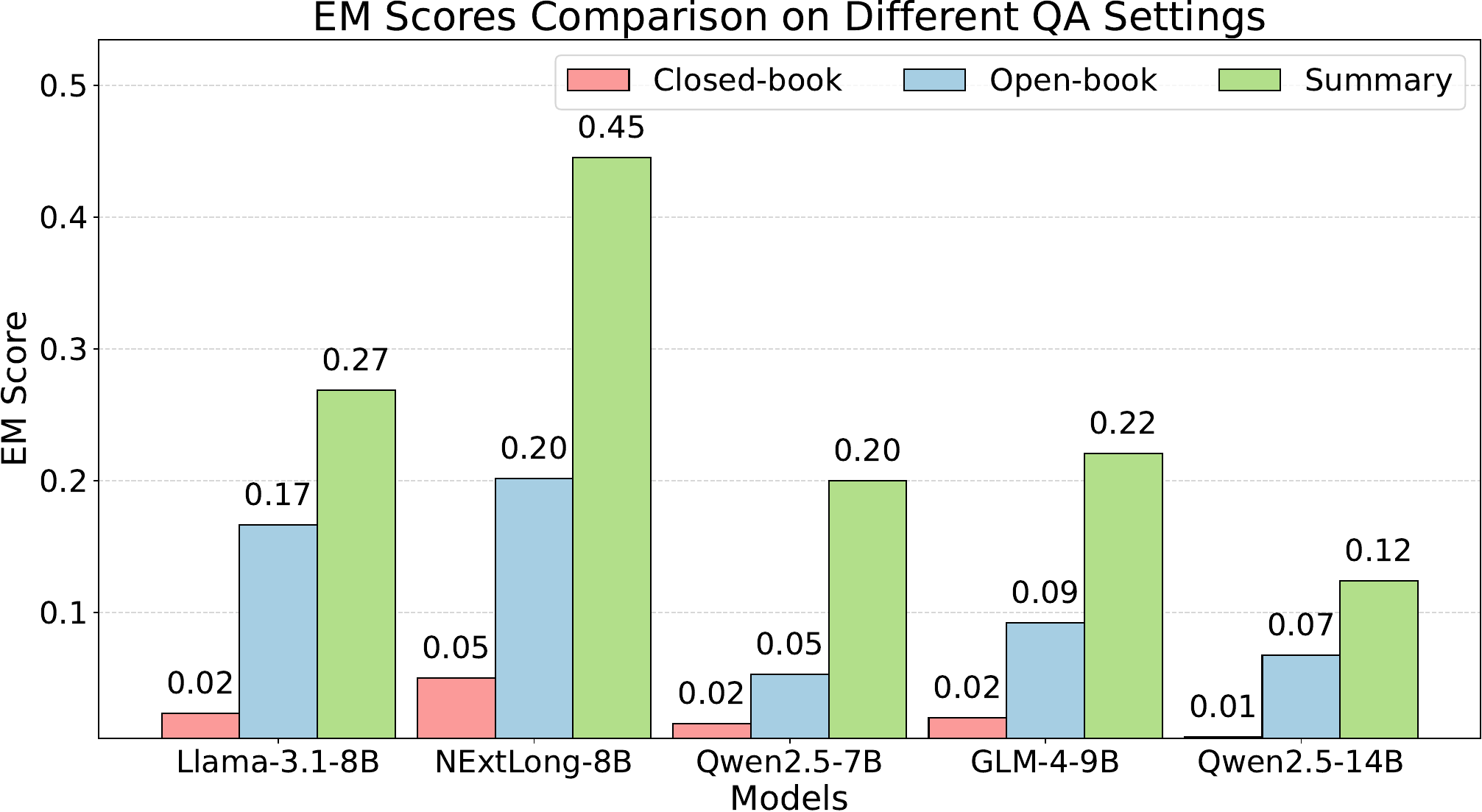}
\caption{EM scores of the models across different settings.}
\label{fig:em_performance_baselines}
\end{figure}

\begin{figure}[h]
\includegraphics[width=\columnwidth]{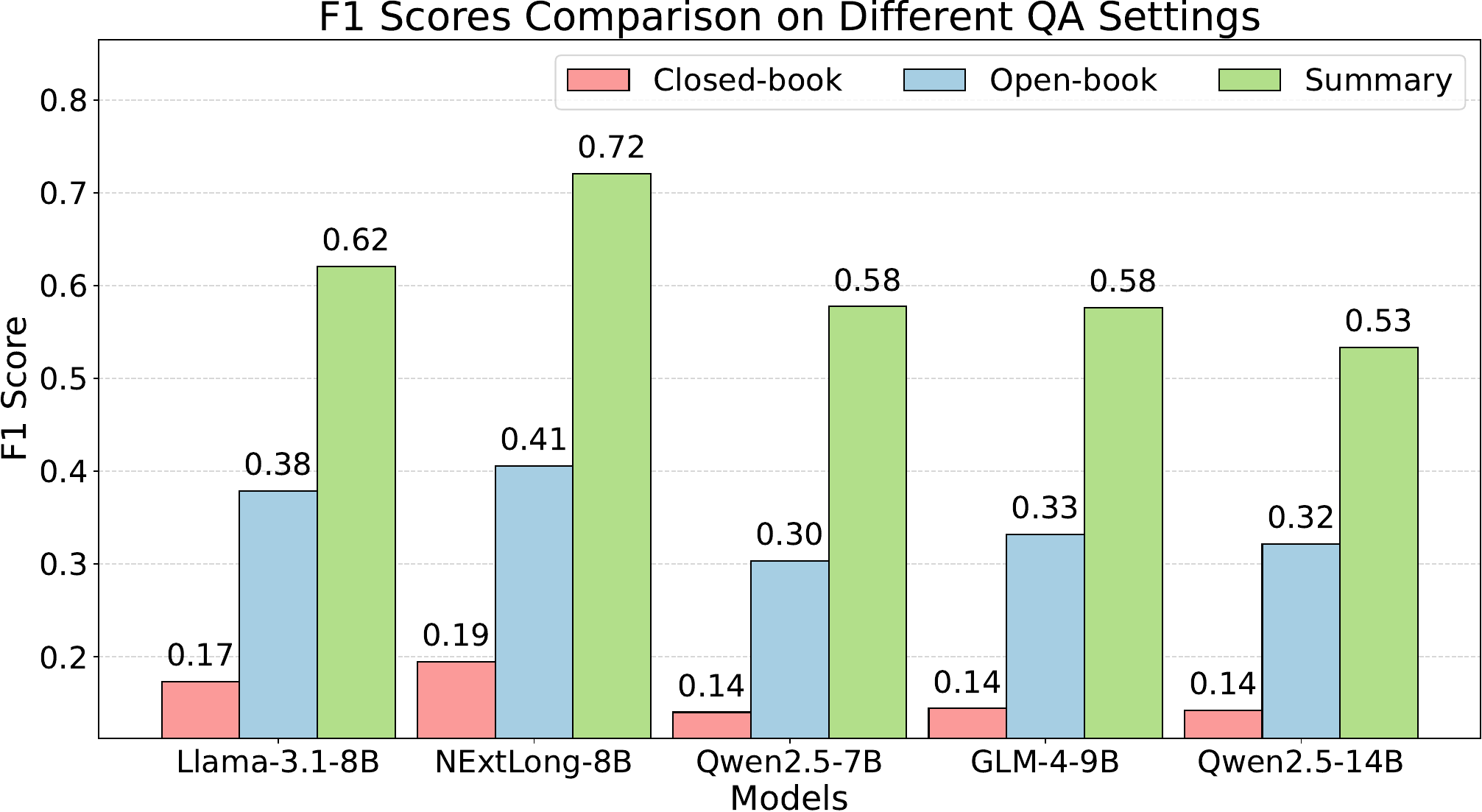}
\caption{F1 scores of the models across different settings.}
\label{fig:f1_performance_baselines}
\end{figure}

\section{Licenses}
We note that NarrativeQA is distributed under the Apache-2.0 License, which permits distributions and modifications.
We adopt the same license when distributing LiteraryQA.
Regarding models, we used closed-sourced options only to evaluate their performance, which complies with their Terms-of-Service (ToS).
The only exception is Claude 3.5 Haiku, which we used through API in our data pipeline.
According to their ToS, this is a legitimate use of their product as we are not developing a competing product and our dataset cannot be classified as harmful.



\begin{table*}[tbp]
\centering
\begin{tabularx}{\linewidth}{|c|X|}
\hline
\textbf{Score} & \textbf{Criteria} \\
\hline
\textbf{1} & The response is completely wrong. \\
\textbf{2} & The output generally deviates from the original question, but there is some information related to the reference answer. \\
\textbf{3} & The response is partially correct, but the generated answer contains some errors, omits key information, or adds \textbf{major extra information} that cannot be validated (in the summary or the references, according to the setting). \\
\textbf{4} & The response is correct \textbf{but} it includes \textbf{minor} details that cannot be verified against the references or summary (according to setting) \\
\textbf{5} & Either exactly the same as one of the references, or a paraphrase of a reference that does not alter its meaning \\
\hline
\end{tabularx}
\caption{Likert Scale Grading Rubric}
\label{tab:grading_rubric}
\end{table*}



\end{document}